\documentclass[conference]{IEEEtran}
\usepackage{glossaries}

\usepackage{eso-pic}
\usepackage{hyperref}
\usepackage{url}

\IEEEoverridecommandlockouts
\usepackage{graphicx}
\usepackage{cite}
\usepackage{float}
\usepackage[table,xcdraw]{xcolor}
\usepackage{amsmath}
\usepackage{hyperref}
\usepackage{cleveref}

\crefname{figure}{Fig.}{Fig.}
\Crefname{figure}{Fig.}{Fig.}

\crefformat{equation}{(#2#1#3)}
\Crefformat{equation}{(#2#1#3)}

\usepackage{gensymb}
\usepackage[T1]{fontenc}    
\usepackage{lmodern}        
\usepackage{csvsimple}
\usepackage{booktabs}
\usepackage{longtable}
\usepackage{csvsimple}
\usepackage{ifthen}

\usepackage[subpreambles=true]{standalone}
\usepackage{import}

\usepackage[compatibility=false]{caption}

\DeclareCaptionFormat{allcaps}{#1#2\uppercase{#3}\par}

\usepackage{subcaption}  

\usepackage{dirtytalk}

\usepackage{acro}
\DeclareAcronym{CNN}{
    short=CNN,
    long=Convolutional Neural Network
}
\DeclareAcronym{COCO}{
    short=COCO,
    long=Common Objects in Context
}
\DeclareAcronym{DSS}{
    short=DSS,
    long=Dynamic Spatial Scaling
}
\DeclareAcronym{FOV}{
    short=FOV,
    long=Field of View
}
\DeclareAcronym{FPS}{
    short=FPS,
    long=Frames Per Second
}
\DeclareAcronym{GPS}{
    short=GPS,
    long=Global Positioning System
}
\DeclareAcronym{IR}{
    short=IR,
    long=Infrared Radiation
}
\DeclareAcronym{OD}{
    short=OD,
    long=Object Detection
}
\DeclareAcronym{SAHI}{
    short=SAHI,
    long=Slicing Aided Hyper Inference
}
\DeclareAcronym{SAR}{
    short=SAR,
    long=Search and Rescue
}
\DeclareAcronym{SOTA}{
    short=SOTA,
    long=State-of-the-Art
}
\DeclareAcronym{UAV}{
    short=UAV,
    long=Unmanned Aerial Vehicle
}
\DeclareAcronym{USV}{
    short=USV,
    long=Unmanned Surface Vessel
}
\DeclareAcronym{mAP}{
    short=mAP,
    long=Mean Average Precision
}
\DeclareAcronym{RGBT}{
    short=RGBT,
    long=RGB and thermal infrared
}
\DeclareAcronym{VAE}{short = VAE, long = Variational Autoencoder}
\DeclareAcronym{SBC}{
    short=SBC,
    long=Single Board Computer
}

\begin{document}

\title{Maritime Small Object Detection from UAVs using Deep Learning with Altitude-Aware Dynamic Tiling}

\author{
\IEEEauthorblockN{Sakib Ahmed\IEEEauthorrefmark{1}, Oscar Pizarro\IEEEauthorrefmark{2}}
\IEEEauthorblockA{\IEEEauthorrefmark{1}Cognitive Neuroinformatics Group, University of Bremen,\\ Bremen, Germany\\
Email: sakib.ahmed@uni-bremen.de}
\IEEEauthorblockA{\IEEEauthorrefmark{2}Department of Marine Technology, NTNU,\\ Trondheim, Norway\\
Email: oscar.pizarro@ntnu.no}
}

\maketitle
\AddToShipoutPictureFG*{%
    \AtPageLowerLeft{%
        \put(0,40){
            \begin{minipage}{\paperwidth}
                \centering
                \footnotesize
                \copyright\ 2025 IEEE. This is the author’s accepted version of an article that has been published by IEEE. \\
                The final published version is available at IEEE Xplore.\\
                \textbf{Full citation:} S. Ahmed and O. Pizarro, “Maritime Small Object Detection from UAVs using Deep Learning with Altitude-Aware Dynamic Tiling,”\\
                \textit{OCEANS 2025 Brest}, BREST, France, 2025, pp. 1-9.
                doi: \href{https://doi.org/10.1109/OCEANS58557.2025.11104659}{10.1109/OCEANS58557.2025.11104659}
            \end{minipage}%
        }%
    }%
}


\begin{abstract}
 \acp{UAV} are crucial in \ac{SAR} missions due to their ability to monitor vast maritime areas.
However, small objects often remain difficult to detect from high altitudes due to low object-to-background pixel ratios.
We propose an altitude-aware dynamic tiling method that scales and adaptively subdivides the image into tiles for enhanced small object detection.
By integrating altitude-dependent scaling with an adaptive tiling factor, we reduce unnecessary computation while maintaining detection performance.
Tested on the SeaDronesSee dataset \cite{dataset_SDSv2} with YOLOv5 \cite{yolov5} and \ac{SAHI} framework \cite{sahi_app}, our approach improves \ac{mAP} for small objects by 38\% compared to a baseline and achieves more than double the inference speed compared to static tiling.
This approach enables more efficient and accurate \ac{UAV}-based \ac{SAR} operations under diverse conditions.
\end{abstract}


\section{Introduction}

Maritime surveillance and \ac{SAR} missions often rely on UAVs to detect and monitor small objects within vast oceanic environments. However, detecting these objects remains challenging due to their limited spatial resolution and the dynamic nature of maritime conditions. While traditional computer vision approaches struggle with the high variability of the environment, \ac{SOTA} deep learning models are constrained by their computational demands and inherent difficulties in detecting small objects, and small and large objects simultaneously \cite{smol_object_anchorbox}.

Existing tiling or slicing strategies can increase small object detection accuracy by examining localized image regions \cite{tiling_power, sahi}.
However, these static tiling approaches incur high computational costs and fail to adapt efficiently across different altitudes and object scales.

In this paper, we propose a novel \textit{Dynamic Tiling Approach}, which integrates altitude-aware spatial scaling (\ac{DSS})with adaptive tiling strategies.
Our pipeline aims to improve detection accuracy and efficiency while respecting the stringent computational constraints of UAV platforms.
By dynamically adjusting the tiling factor in response to operational conditions and stages, our approach maintains high detection performance across varying object scales and distances.

The remainder of this paper is structured as follows: Section II briefly reviews related works. Section III presents the proposed altitude-aware dynamic tiling method. Experimental procedures are outlined in Section IV, followed by results and analysis in Section V. Finally, conclusions and future directions are presented in Section VI.
\section{Related Works}

Several studies have investigated maritime object detection using UAVs. Zolich et al. \cite{Zolich2021-tk} presented a \ac{UAV}-USV (Unmanned Surface Vessel) system for object recovery using active payload markers (Aruco Tags, \ac{GPS}), which requires \textit{active} targets and favorable conditions, thus limiting general applicability. Classical image processing approaches using thermal cameras and filters \cite{Leira2021-mo, Leira_2015_detect, Leira_2015_georef} have demonstrated effectiveness under specific conditions but often lack generalization, particularly under variable maritime environments, hence not scalable.

Recent approaches employing \acp{CNN} have substantially improved detection accuracy and generalization. Li et al. \cite{Li_Wu_Kittler_2018} demonstrated that fusing thermal and visible imagery using deep learning achieves superior object detection performance. Kumar et al. \cite{Kumar2022ImprovedYOLO3} specifically addressed small object detection in UAV imagery by modifying YOLOv3, demonstrating that CNN-based single-stage detectors can be both efficient and accurate for UAV-based maritime detection tasks.

\ac{RGBT} i.e., visible-thermal sensor fusion has become a popular strategy to overcome visibility challenges. Models like MFGNet \cite{mfgnet_rgbt} utilize dynamic modality-aware filtering to improve robustness to changing environmental conditions. While promising, such methods typically rely heavily on extensive multimodal data, and misalignment between RGB and thermal images often remains challenging \cite{rgbt-alignment}.

To address dataset limitations, synthetic data generation and semi-supervised learning have emerged as alternatives. Approaches like synthetic \ac{IR} image generation using CycleGAN and \acp{VAE} for anomaly detection \cite{Drone_rgbir, rgbt_salient} have shown potential but also limitations due to insufficient realism or excessive computational complexity.

Walker et al. \cite{oscar_physics_scaling} conducted research on physics-based image correction (scaling) for the benthic image segmentation task and showed performance gain on transfer learning. They also showed that fidelity loss due to image scaling has fewer performance penalties compared to the performance gain achieved by the scale correction. This is especially interesting for us since we will also explore the premise of sacrificing fidelity to reduce computational overhead in our approach.

In summary, while existing methods offer solutions under specific constraints, they either require instrumented targets, struggle with environmental generalization, or demand substantial computational resources. This motivates our altitude-aware dynamic tiling approach, which uniquely integrates altitude-dependent scaling (\ac{DSS}) and adaptive tile generation, ensuring robust and computationally efficient detection of maritime small objects across varying operational conditions.


\section{Proposed Methodology}

\subsection{Adaptive Altitude-Aware Scaling}
To determine the optimal image resolution $\hat{p}$ required for reliable object detection, we employ the relationship derived in \cite{holistic_drone}:
\begin{equation}
\hat{p} = 2h \cdot \tan\left(\frac{\hat{\phi}}{2}\right) \cdot \sqrt{\frac{rec}{obj}}
\label{eq:opt_res}
\end{equation}
Here, $h$ is the altitude, $\hat{\phi}$ is the camera \ac{FOV}, $rec$ is the minimum pixel area required for recognition, and $obj$ is the target object size (or the smallest of the target classes).

This scaling ensures that even small objects meet recognition criteria at different flight altitudes.


\subsection{Tiling}

\begin{figure}[hbtp!]
    \centering
    \includegraphics[width=0.48\textwidth]{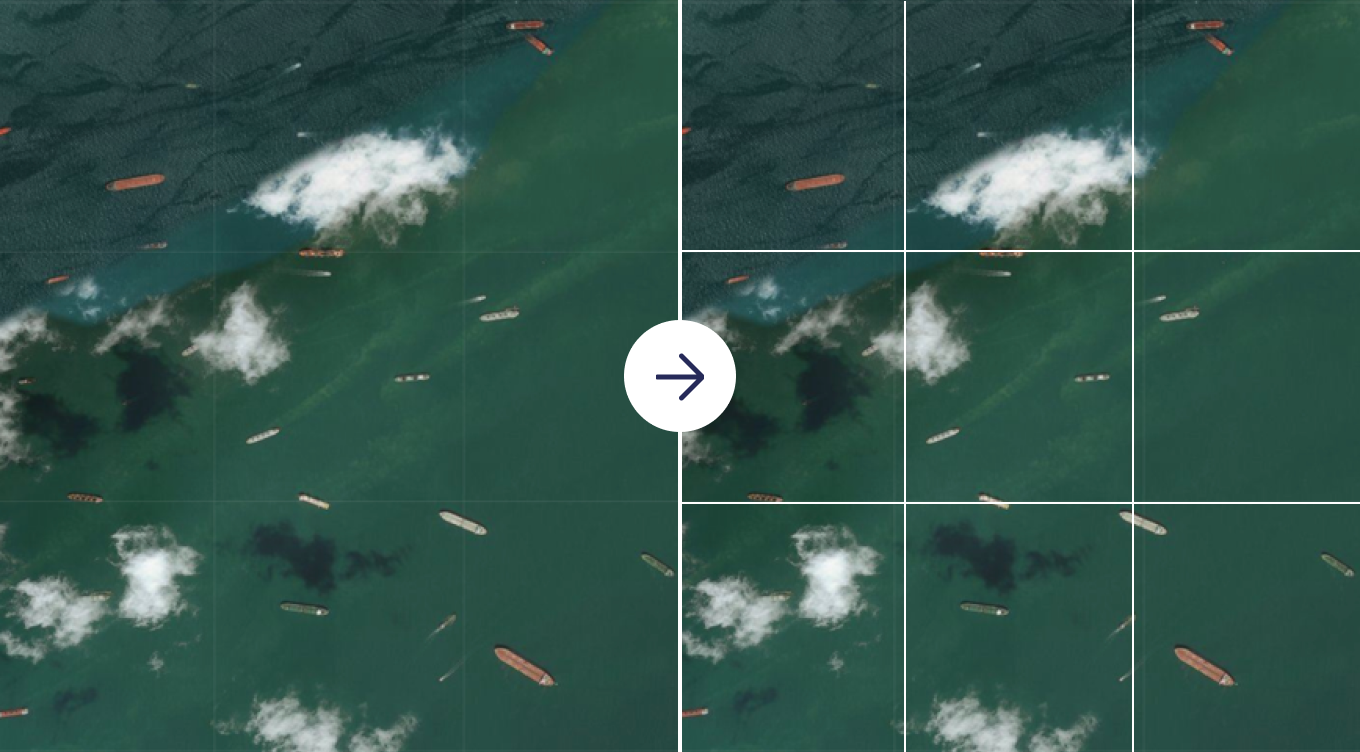}
    \caption{Illustration of a general tiling process (without overlap), adapted from \cite{Raster_Tiling}.}
    \label{fig:tilingggg}
\end{figure}

Tiling is a generic preprocessing block that clips rasters into rectangle-shaped tiles. Tiling is especially important for large raster datasets like satellite or aerial images that should be broken into more manageable pieces to fit into memory and improve performance. Typically, overlapping tiles are used to avoid missing or clipped objects at the edges, as we can see happens in \Cref{fig:tilingggg}.

\subsection{How \ac{SAHI} works}
The \ac{SAHI} framework divides an input image into
${N}\times{N}$ overlapping patches(tiles).
Object detection is then performed on each tile separately. 
The predictions from the overlapping tiles are merged back into the original image space using Non-Maximum Suppression (NMS).
During NMS, bounding boxes with Intersection-over-Union (IoU) ratios above a specified threshold $T_m$ are consolidated,  while detections with a confidence score below a threshold $T_d$ are discarded \cite{sahi}.

The authors of \ac{SAHI} also maintained that smaller patch sizes increase the likelihood that larger objects will not fit completely within a single patch, thus negatively impacting detection accuracy \cite{sahi}.
Interpreted in our application context, when a UAV decreases altitude (such as during the recovery phase toward a floating asset), the object's apparent \textit{pixel-print} enlarges. Consequently, employing a fixed tiling strategy risks degrading detection performance due to the clipping effect.


\subsection{Proposed Dynamic Tiling for Multiscale Detection}
To mitigate \ac{SAHI}'s limitations, we propose an altitude-aware \ac{DSS} integrated with adaptive tiling, leveraging the \ac{SAHI} framework.
After scaling the images according to altitude  (as per \eqref{eq:opt_res}),
we apply the \ac{SAHI} framework to slice each scaled image into overlapping tiles.
As part of this pipeline, a YOLOv5s model (though any supported model may be used) trained on tiled images processes each tile individually.
The key advantage is that the tiling factor is dynamically adjusted based on the altitude-dependent image scaling (\ac{DSS}).
This adaptive approach reduces computational overhead at lower altitudes and ensures that objects appear at suitable scales for better anchor box matching \cite{liu2016ssd}.

Increased tiling reduces \ac{FPS} due to more patches per frame, but at higher altitudes the target stays in view longer \cite{holistic_drone}, offsetting lower \ac{FPS}.
Conversely, fewer tiles are necessary at lower altitudes. This is because the effective spatial resolution is already sufficiently high, improving \ac{FPS} when speed is most critical. Thus, the optimal number of tiles per dimension, ${n}$, implicitly depends on $altitude$, camera ${FOV}$, and the target $object \;size$:

\begin{equation} \label{eq:tile_func}
n = f(altitude, FOV, object\; size)
\end{equation}

\begin{figure}[hbpt!]
    \centering
    \includegraphics[width=0.48\textwidth]{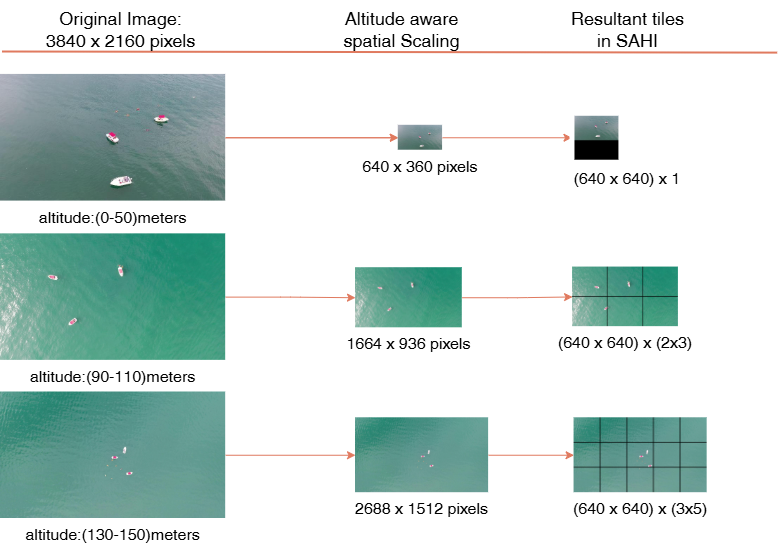}
    \caption{Dynamic tiling workflow: altitude-aware scaling (\ac{DSS}) followed by tiling to facilitate object detection.}
    \label{fig:tiling}
\end{figure}

\begin{figure}[hbpt!]
    \centering
    \includegraphics[width=0.48\textwidth]{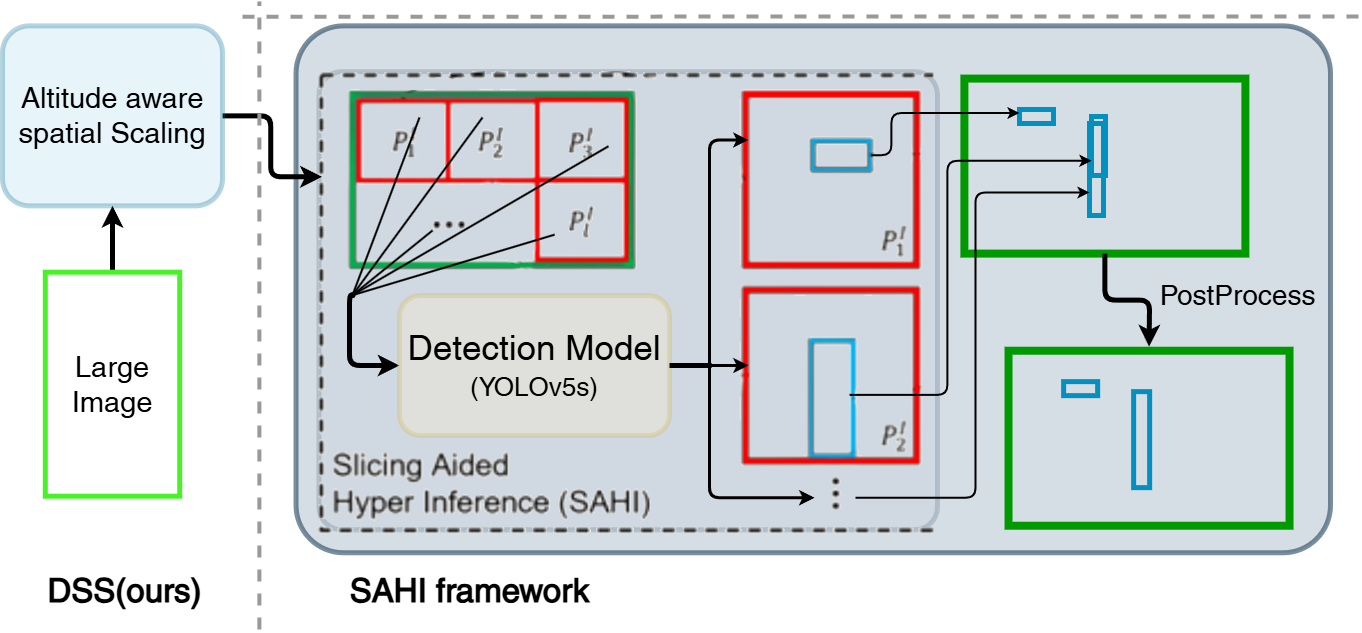}
    \caption{\ac{DSS} integrated with the \ac{SAHI} framework and YOLOv5.}
    \label{fig:framework}
\end{figure}

Practically, ${n}$ is not explicitly computed separately; instead, it emerges naturally from altitude-aware scaling (as per \eqref{eq:opt_res}) combined with the fixed tile size(640x640 pixels) we used within the \ac{SAHI} framework. 
\Cref{fig:tiling} and \Cref{fig:framework} illustrate the overall workflow and system architecture.

\section{Experimental Setup}

\subsection{Dataset}
For experimenting and development, we selected the \textbf{\textit{SeaDronesSee}} Dataset \cite{dataset_SDSv2}. It is the first large-scale annotated (including camera metadata like altitude and gimbal angles/camera orientations ) UAV-based dataset of swimmers in open waters. The class labels are swimmers (swimmers with and without life jackets), life-saving appliances, boat(speed boat), buoy(green and red),  and jet-ski.  The dataset offers several challenges, i.e.,  object detection,  single-object tracking, and multi-object tracking. We only used the \textbf{\textit{Object Detection v2}} for our purposes. This set contains around 8.9K Training, 1.5K validation, and 3.7K test images and annotations in COCO-json format \cite{cocodataset}.

\begin{figure}[hbpt!]
        \centering
        \begin{subfigure}[b]{0.28\textwidth}
            \centering
            \includegraphics[trim=0 0 0 0\textheight,clip,width=0.94\textwidth]{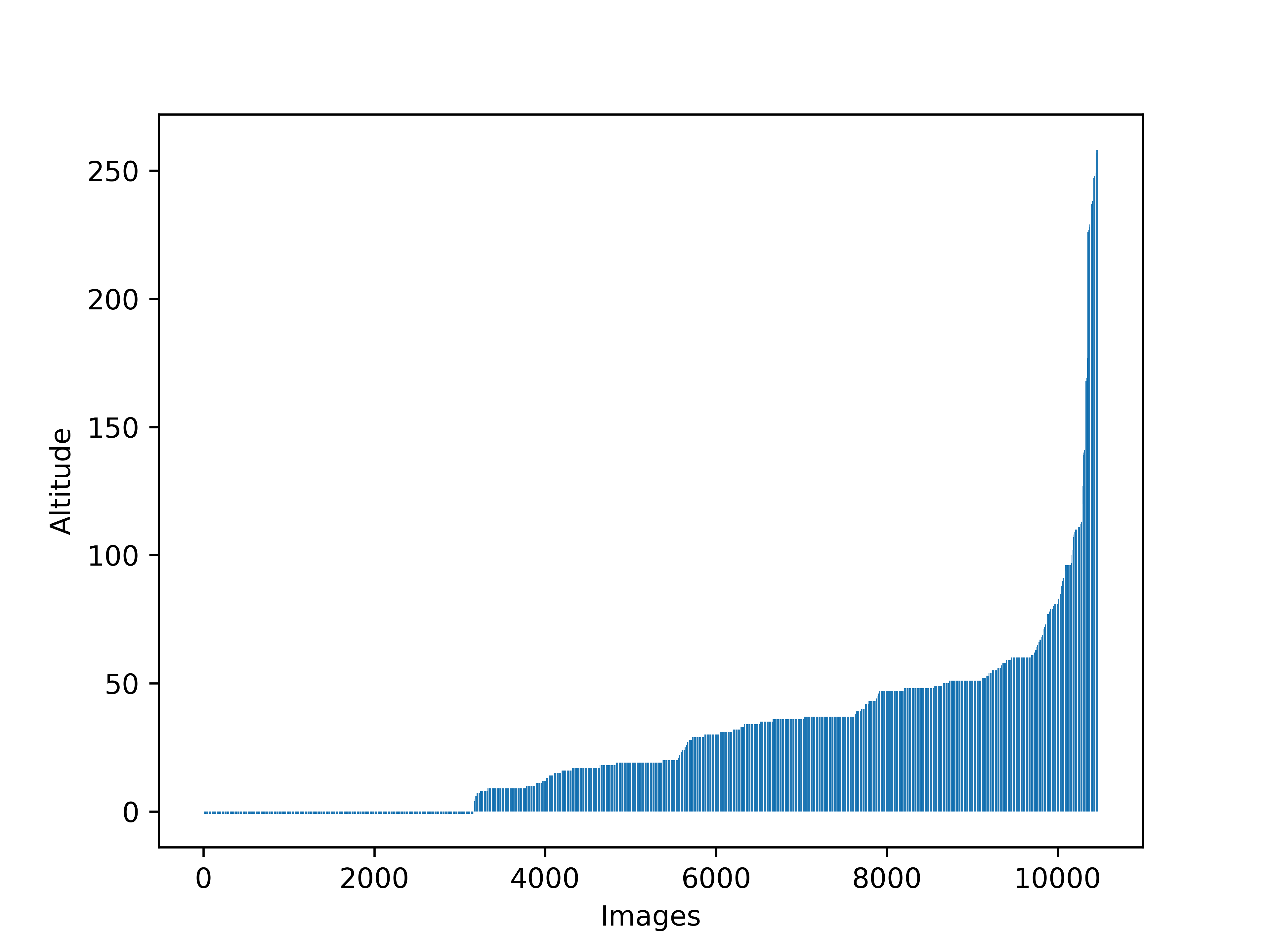}
            \caption{Altitude 0 indicates missing altitude information.}
		\label{sfig:all_at}
        \end{subfigure}
        \begin{subfigure}[b]{0.20\textwidth}
            \centering
            \includegraphics[trim=0 0 0 0\textheight,clip,width=0.99\textwidth]{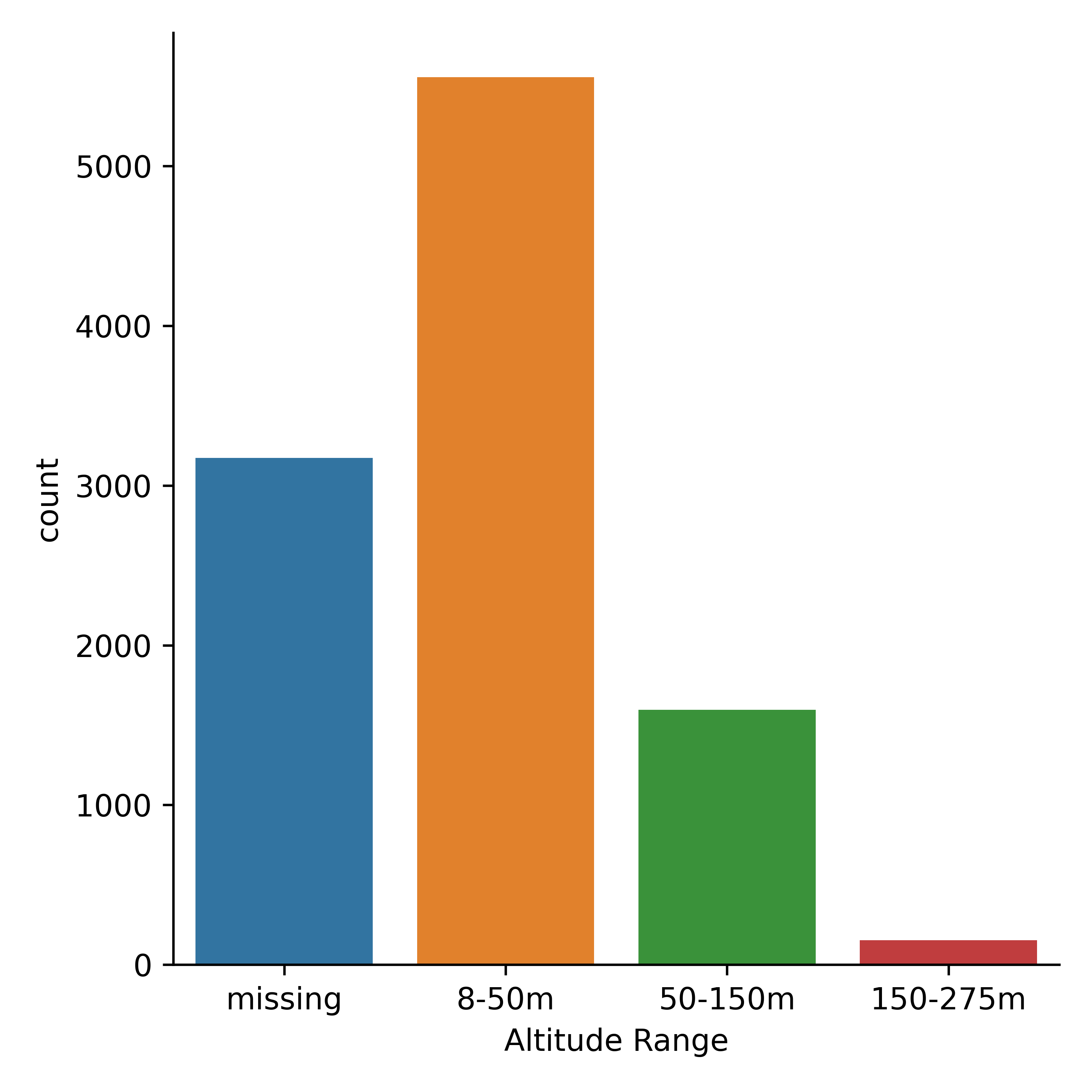}
            \caption{Dataset's image distribution altitude-wise.}
    		\label{sfig:DDDDDDDDDDDDDDDDDDD}
        \end{subfigure}
            
        \caption {Dataset Image Altitudes.}
        \label{fig:imageAltitudes}
    \end{figure}

The authors have a brief description on their website about the data acquisition process. We got further insight from the annotation files, which also contain image metadata for most of the images. Presumably, the images were captured using a DJI Mavic series quadcopter (4k resolution image frames extracted from video) from altitudes ranging between 9--140 meters, and a Trinity F90 series fixed-wing drone which produced (in the training and validation sets) two types (mostly high altitude) images: \textbf{i.} 3172 images of $5456\times3632$ pixels (20M) resolution without altitude information; \textbf{ii.} about 500 RGB images from a multi-spectral camera of size around $\sim 1230 \times 930$ pixels (altitude range from 20--259 meters, but mostly 200+). See \Cref{fig:imageAltitudes}.

Upon careful visual inspection of apparent object size(dimensions in pixels)  of common objects(i.e., same boats),  and comparing them between other images in the dataset(that includes the altitude data), using
simplified pinhole camera model \cite{camera_dist}, we estimated that the 20M images with missing altitudes are mostly taken from altitudes beyond 200 meters.


\subsection{Dataset Preprocessing}
\label{sec:dt_preprocess}

\begin{figure}[!hbtp]
    \centering
    \begin{subfigure}[b]{0.23\textwidth}
        \centering
        \includegraphics[trim=0 0 0 0\textheight,clip,width=.99\textwidth]{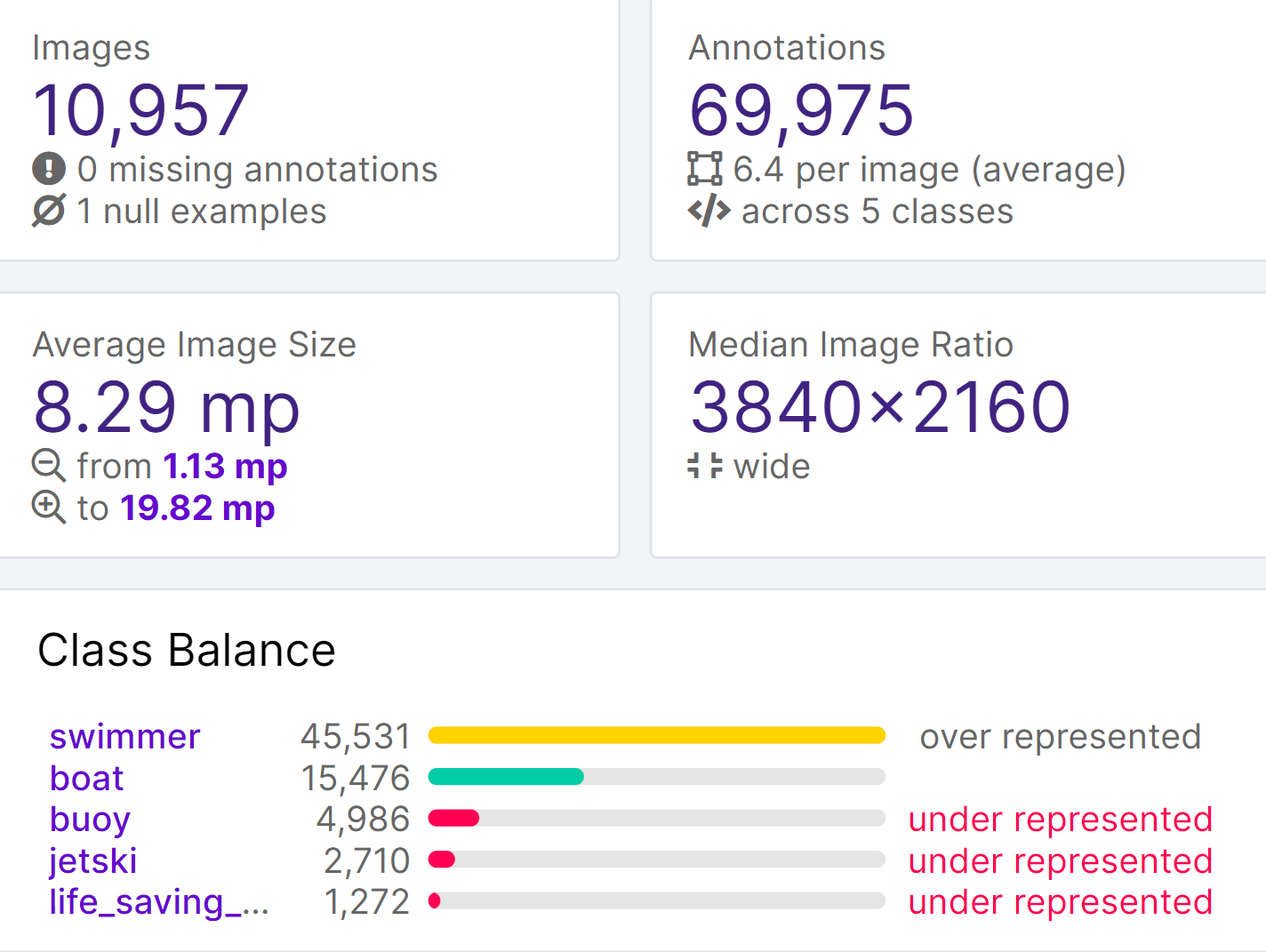}
        \caption{ Dataset Overview}
        \label{sfig:dataset_overview}
    \end{subfigure} 
    \begin{subfigure}[b]{0.20\textwidth}
        \centering
        \includegraphics[trim=0 0 0 0\textheight,clip,width=.99\textwidth]{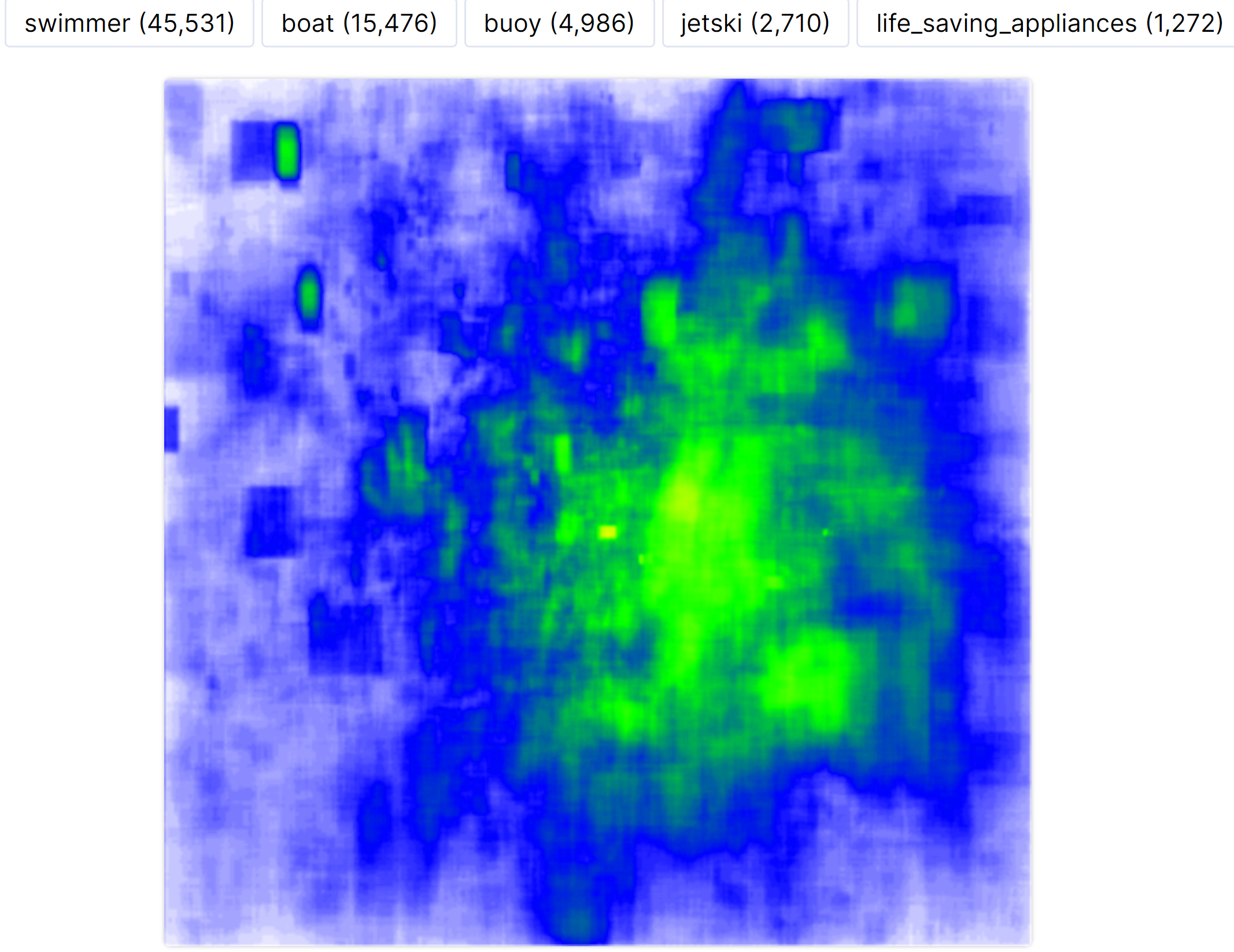}
        \caption {Annotation Heatmap (all classes)}
        \label{sfig:Annotationheatmap}
    \end{subfigure}
    \hfill

    \begin{subfigure}[b]{0.23\textwidth}
        \centering
        \includegraphics[trim=0 0 0 0\textheight,clip,width=.99\textwidth]{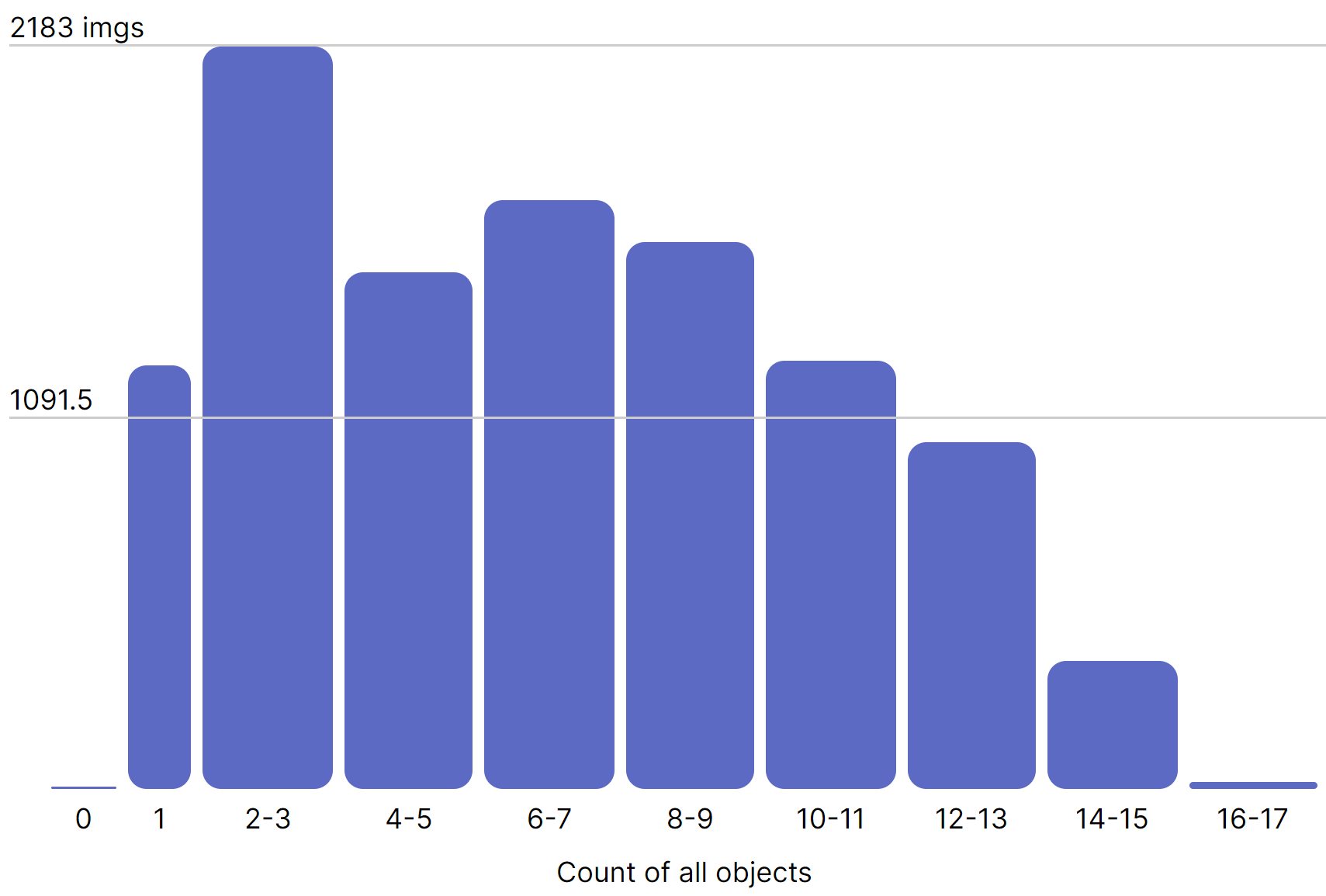}
        \caption{ Histogram of Object Count by Image}
        \label{sfig:ayuffyufyufufutyftuyfufuytftf}
    \end{subfigure}
    \begin{subfigure}[b]{0.23\textwidth}
        \centering
        \includegraphics[trim=0 0 0 0\textheight,clip,width=.99\textwidth]{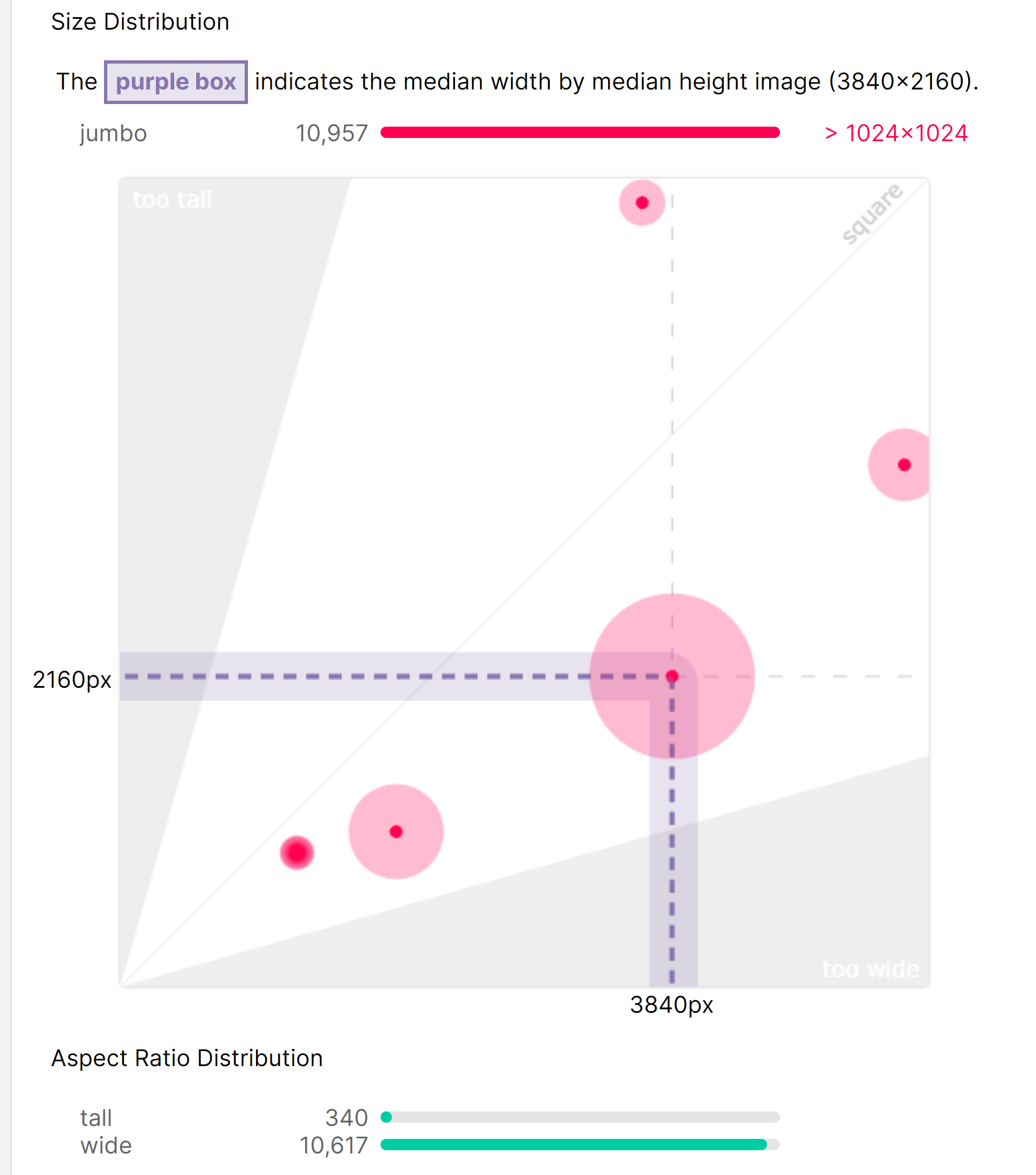}
        \caption{ Image dimensions}
        \label{sfig:imagedims}
    \end{subfigure}
    
    \caption{Dataset Insights (generated by \cite{roboflow}).}
    \label{fig:datasetInsight}
\end{figure}

As the ground truth annotations for the \textit{test set} are withheld by the organizers, we skipped the original \textit{test set}, merged the train and validation sets, and created a new train-val-test split (70\%-20\%-10\%) using Roboflow \cite{roboflow}.  
A detailed overview of the dataset is shown in \Cref{fig:datasetInsight}.

\subsubsection{Training sets Preprocessing.}
\label{sec:yolo-trainset}

\begin{figure}[!hbpt]
    \centering
    \includegraphics[width=.45\textwidth]{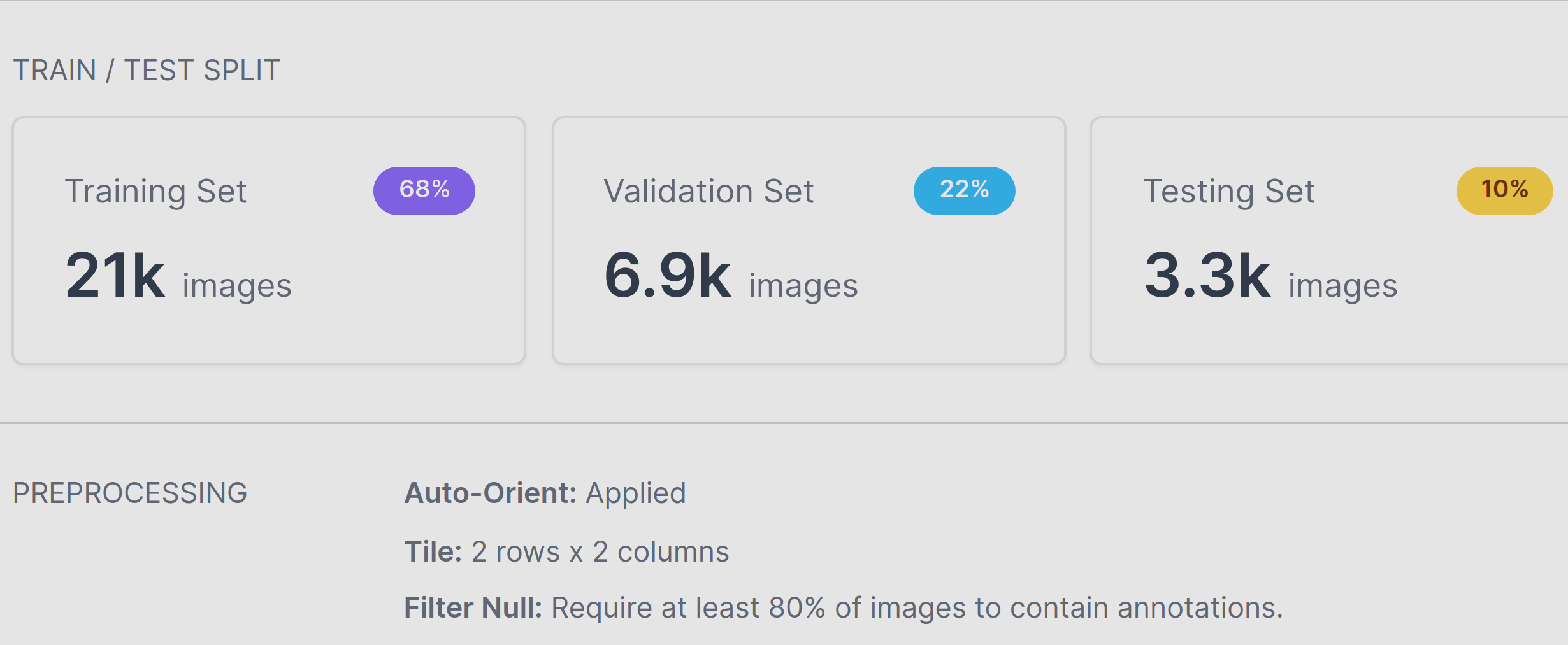}
    \caption{Preprocessing of Training sets.}    \label{fig:yolo_train_val_preprocess}
\end{figure}

For the training and validation sets, the images were tiled in $4(2\times2)$ images without applying any other augmentations using Roboflow. Then we filtered the generated images so that at least 80\% of the image segments contain objects.
\Cref{fig:yolo_train_val_preprocess} shows the result of the described steps. Note that this process also generated tiled test sets, which were discarded.

\subsubsection{Test set Preprocessing}
First, we generate \Cref{tab:testsetscalingres} using \eqref{eq:opt_res}, for an object of size, 0.48 $m^2$ (typical human size, 0.3m x 1.6m ) and considering the detection criterion to a minimum of 80 $pixel^2$ as suggested on \cite{holistic_drone}. 
By setting $rec=80\,\text{pixels}^2$ and $obj=0.48\,\text{m}^2$, we obtain altitude-dependent minimum image sizes that would still retain the fidelity for detection. 

\begin{table}[hbpt!]
\centering
\caption{Minimum required image dimension at different altitudes}
\label{tab:testsetscalingres}
\begin{tabular}{@{}llll@{}}
\toprule
Altitude & $\hat{p}$       & $\hat{p}$ (DJI) & $\hat{p}$ (trinity) \\ 
(meters) & (trinity 20M)   & (MAVIC 4K)      & (Multispectral RGB)  \\
         & FOV 73$\degree$ & FOV 65$\degree$ & FOV 50$\degree$     \\ \midrule
10       & 191             & 164             & 120                  \\
20       & 382             & 328             & 240                 \\
30       & 573             & 493             & 361                 \\
.        & .               & .               & .                    \\
.        & .               & .               & .                    \\
240      & 4585            & 3947            & 2889                \\
250      & 4776            & 4112            & 3010                \\ \bottomrule
\end{tabular}
\end{table}

Since our dataset is not uniform and contains images of different sizes and qualities from multiple cameras, to simplify practical implementation, we scale the original images (of our test set) into six discrete resolution \say{bins}. These \say{bins} are discrete approximations of \(\hat{p}\) for practical implementation. This is our \ac{DSS} step adopted for this particular dataset.
The lowest bin $\bar{P}_1$ was 640 pixels. Each subsequent bin \(\bar{P}_i\) increases the scaled image dimension by $80\%$ of the base tile size (640 pixels)\eqref{eq:res_bins}. This enforces a tiling stride of 80\%, or in other words, tiles with $20\%$ overlap in each dimension.
\begin{equation}
\bar{P}_{i} = {N}+{N}\times{stride}\times({i}-1) 
\label{eq:res_bins}
\end{equation}

Where ${N}$ is the tile size and \(i\) denotes the bin number (\(i=1,2,\dots,6\)).
Higher-resolution images are scaled down to the nearest bin, ensuring resolutions remain above \(\hat{p}\) where possible (e.g., an image at \(60\,m\) \(\rightarrow\) \(\hat{p}=986\,px\) \(\approx\) bin \(\bar{P}_2\), \(1152\,px\)). However, images with resolutions below \(\bar{P}_1\) (640px) are not up-scaled, to avoid interpolation artifacts, potential degradation in detection performance, and unnecessary computational overhead.

Two scaled test sets, ScaledV1 and ScaledV2, were generated to capture variations in altitude and resolution, as detailed in \Cref{tab:testsets} using ${N}=640$ (matching the default input size of YOLOv5) and ${stride=80\%}$.
The ScaledV1 policy was relatively conservative; it downscaled images at 640px up to 50m altitude, then stepped up to higher resolutions for higher altitudes, whereas ScaledV2 was more aggressive, starting to increase resolution at a lower altitude threshold (around 30m).
In general, both ScaledV1 and ScaledV2 ensure that high-altitude images are scaled down to a manageable size ( without compromising detection fidelity significantly beyond the requirements dictated by \eqref{eq:opt_res}).

\begin{table}[hbpt!]
    \centering
\caption{Details of the two test sets (ScaledV1 and ScaledV2) created after applying \ac{DSS}}
    \resizebox{\columnwidth}{!}{%
    \begin{tabular}{llllll} 
        \toprule
        \begin{filecontents*}{test-dataset-scaling1.csv}
Image source,Image Size,Altitude,ScaledV1,Altitude,ScaledV2
,(pixel),(meter),(pixel),(meter),(pixel)
Trinity,1330x930,0-50,640x480,,
Multispectral,,50-150,1152x864,0-150,1152x864
RGB,,150+,1664x1248,150+,1664x1248
,,,,,
DJI Mavic,3840x2160,0-50,640x360,0-30,640x360
4K,,50-90,1152x648,30-60,1152x648
,,90-110,1664x936,60-90,1664x936
,,110-130,2176x1224,90-120,2176x1224
,,130-150,2688x1512,120-150,2688x1512
,,,,,
Trinity (20M),5456x3632,missing,3712x2470,missing,4224x2816
\end{filecontents*}
\csvreader[
            head=false,
            late after line=\\,
            filter equal={\thecsvinputline}{1} 
        ]{test-dataset-scaling1.csv}{}{
            \csvcoli & \csvcolii & \csvcoliii & \csvcoliv & \csvcolv & \csvcolvi
        }

        \begin{filecontents*}{test-dataset-scaling.csv}
Image source,Image Size,Altitude,Scaled V1,Altitude,Scaled V2
,(pixel),(meter),(pixel),(meter),(pixel)
Trinity,1330x930,0-50,640x480,,
Multispectral,,50-150,1152x864,0-150,1152x864
RGB,,150+,1664x1248,150+,1664x1248
,,,,,
DJI Mavic,3840x2160,0-50,640x360,0-30,640x360
4K,,50-90,1152x648,30-60,1152x648
,,90-110,1664x936,60-90,1664x936
,,110-130,2176x1224,90-120,2176x1224
,,130-150,2688x1512,120-150,2688x1512
,,,,,
Trinity (20M),5456x3632,missing,3712x2470,missing,4224x2816
\end{filecontents*}
\csvreader[
            head=false,
            late after line=\\,
            filter equal={\thecsvinputline}{2} 
        ]{test-dataset-scaling.csv}{}{
            \csvcoli & \csvcolii & \csvcoliii & \csvcoliv & \csvcolv & \csvcolvi
        }

        \midrule

        \csvreader[
            head=false,
            late after line=\\,
            filter test={\ifnumgreater{\thecsvinputline}{2}} 
        ]{test-dataset-scaling.csv}{}{
            \csvcoli & \csvcolii & \csvcoliii & \csvcoliv & \csvcolv & \csvcolvi
        }
        
        \bottomrule
    \end{tabular}%
    }
\label{tab:testsets}
\end{table}

\begin{figure*}[hbtp!]
    \centering
    \begin{subfigure}[b]{0.32\textwidth}
        \centering
        \includegraphics[width=\textwidth]{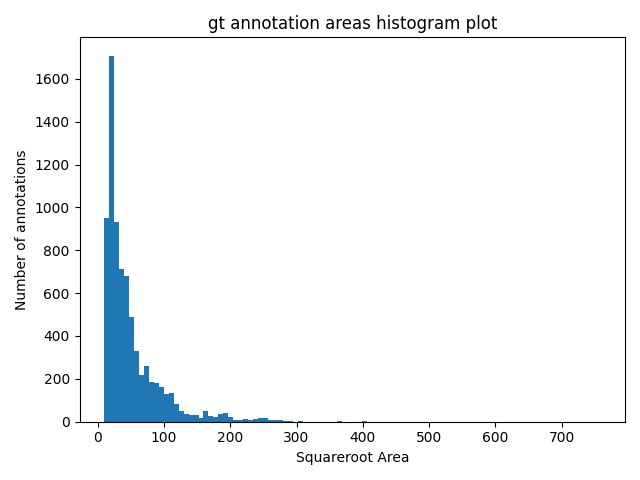}
    \end{subfigure}
    \hfill
    \begin{subfigure}[b]{0.32\textwidth}
        \centering
        \includegraphics[width=\textwidth]{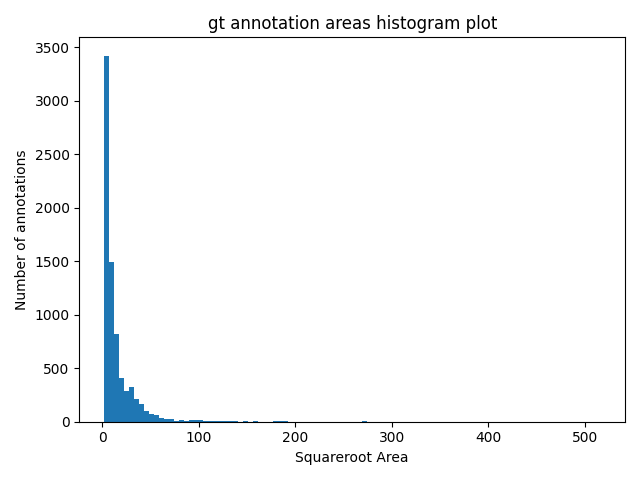}
    \end{subfigure}
    \hfill
    \begin{subfigure}[b]{0.32\textwidth}
        \centering
        \includegraphics[width=\textwidth]{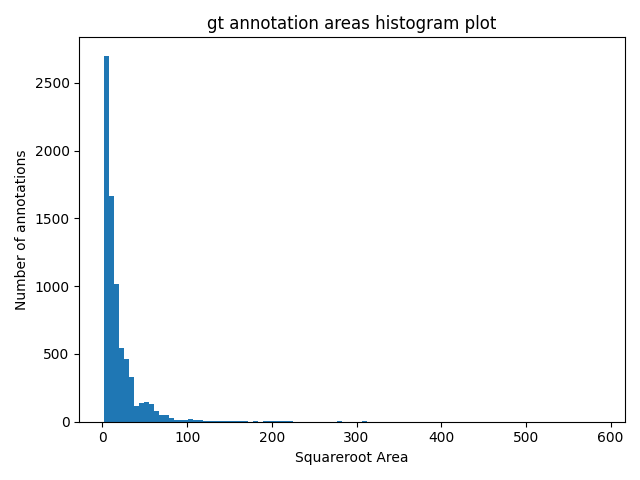}
    \end{subfigure}

    \medskip 

    \begin{subfigure}[b]{0.32\textwidth}
        \centering
        \includegraphics[width=\textwidth]{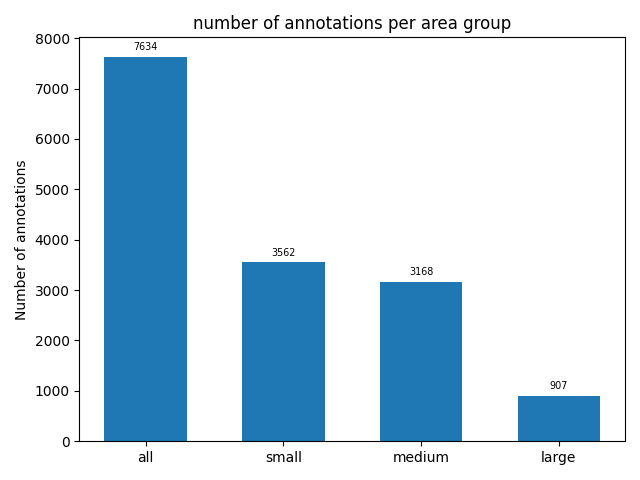}
        \caption{Original}
        \label{sfig:og_ann_area_hist}
    \end{subfigure}
    \hfill
    \begin{subfigure}[b]{0.32\textwidth}
        \centering
        \includegraphics[width=\textwidth]{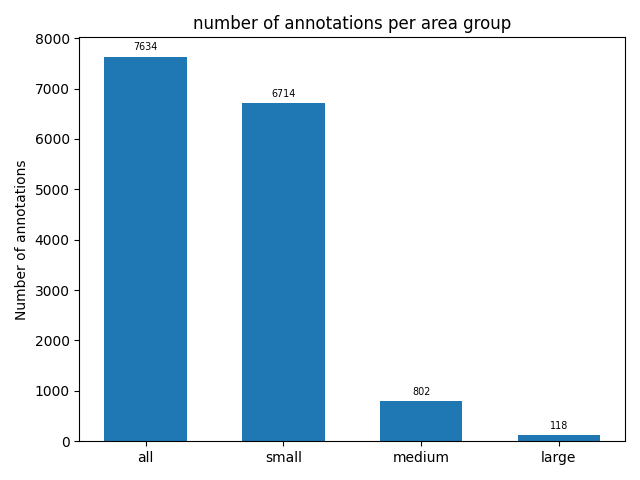}
        \caption{ScaledV1}
    \end{subfigure}
    \hfill
    \begin{subfigure}[b]{0.32\textwidth}
        \centering
        \includegraphics[width=\textwidth]{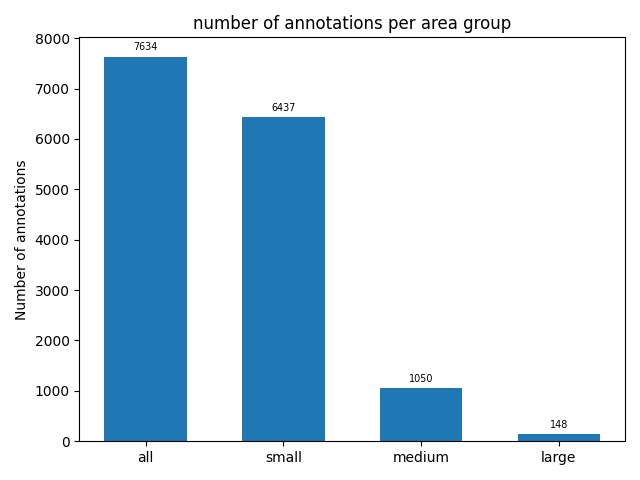}
        \caption{ScaledV2}
    \end{subfigure}

    \caption{Ground-truth (gt) annotation distribution for the test set in terms of pixel area. The top row shows histograms of annotation areas, and the bottom row shows the number of annotations per area group.}
    \label{fig:testsets}
\end{figure*}

\Cref{fig:testsets} shows the object area(pixel-print) comparison between the original and the two scaled versions of the test set. From \Cref{fig:datasetInsight}, we know that most of our annotated objects are from the \textit{swimmer} class, hence small objects. However, from the annotation histogram in \Cref{sfig:og_ann_area_hist}, we can see that around 40\% of the annotations in the original dataset (test set) are in the medium group. This indicates a high scale variance of apparent object sizes in the pictures.
After scaling (\ac{DSS}), we can see that most of the objects are now small objects, and the \textit{small-medium-large} distribution did not change much between our two test sets, and the \textit{pixel-print} distribution is more in line with the physical size of the objects.


\subsection{Model Selection}
\label{sec:dt_model_selection}
We selected YOLOv5 \cite{yolov5} as our object detection model for this experiment. This could be done using any of the \ac{SOTA} object detection models, but we decided to use YOLOv5 as it was one of the best \ac{OD} models at the time of the experiments. It is still actively maintained, easy to implement, supports a variety of \ac{OD} tasks, including Bounding box detection, classification, and segmentation, and most importantly, has built-in support for \acp{SBC} like Raspberry Pi and NVIDIA Jetson for field deployment in the future.

\subsection{Training}
\label{sec:dt_training}

Consequently, we are using the \say{s} (small) variant of the YOLOv5 model, with the official Checkpoints (predefined weights with 300 epochs on COCO dataset) named \say{YOLOv5s}.

\begin{figure*}[htbp!]
    \centering
    \begin{subfigure}[b]{0.55\textwidth}
        \centering
        \includegraphics[width=\textwidth]{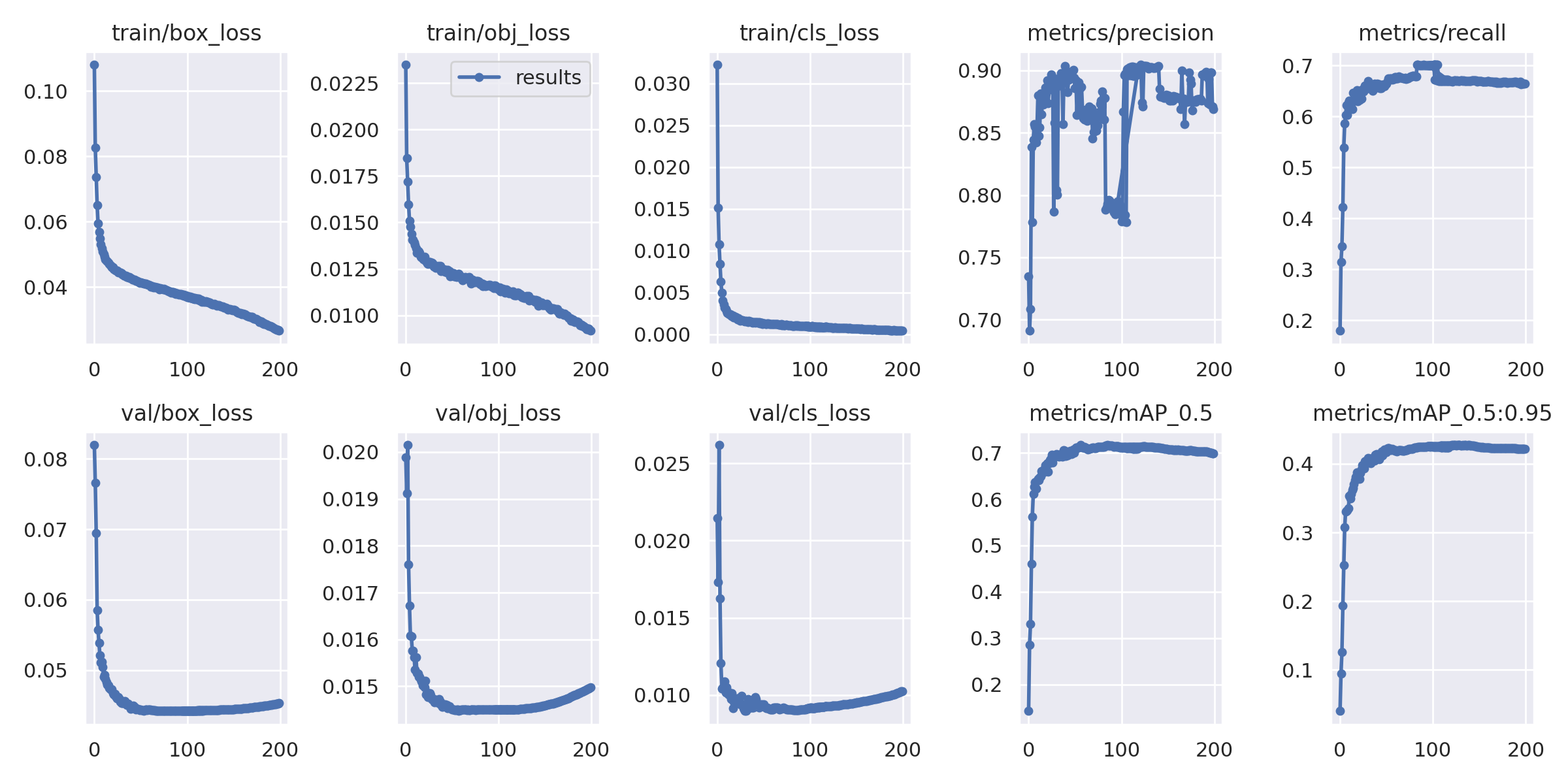}
        \caption{Training epochs}
        \label{sfig:yolotraininggraph}
    \end{subfigure}
    \hfill
    \begin{subfigure}[b]{0.41\textwidth}
        \centering
        \includegraphics[width=\textwidth]{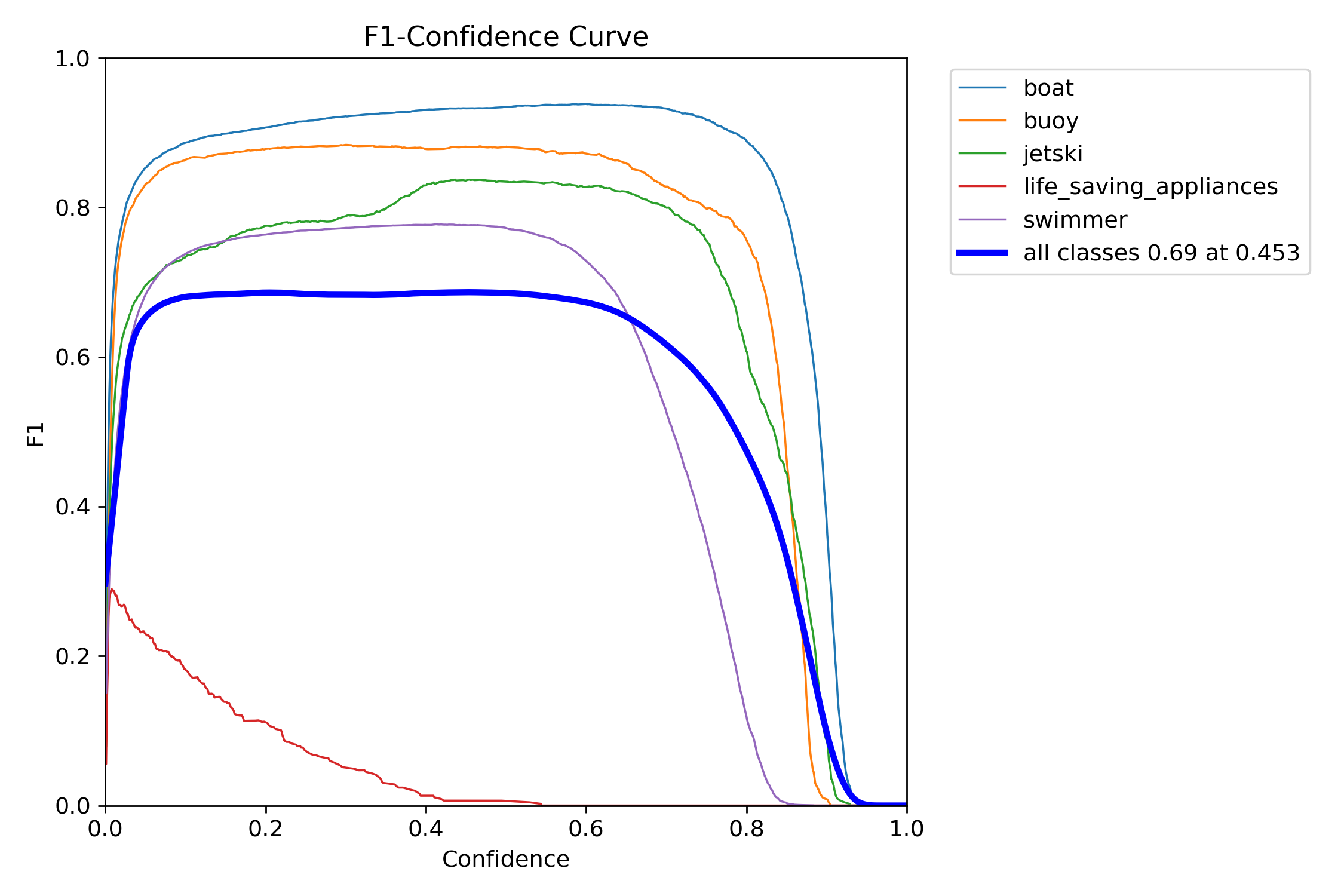}
        \caption{F1-confidence (training)}
        \label{sfig:yolotrainf1}
    \end{subfigure}
    \hfill
    \medskip 
    \begin{subfigure}[b]{0.48\textwidth}
        \centering
        \includegraphics[width=\textwidth]{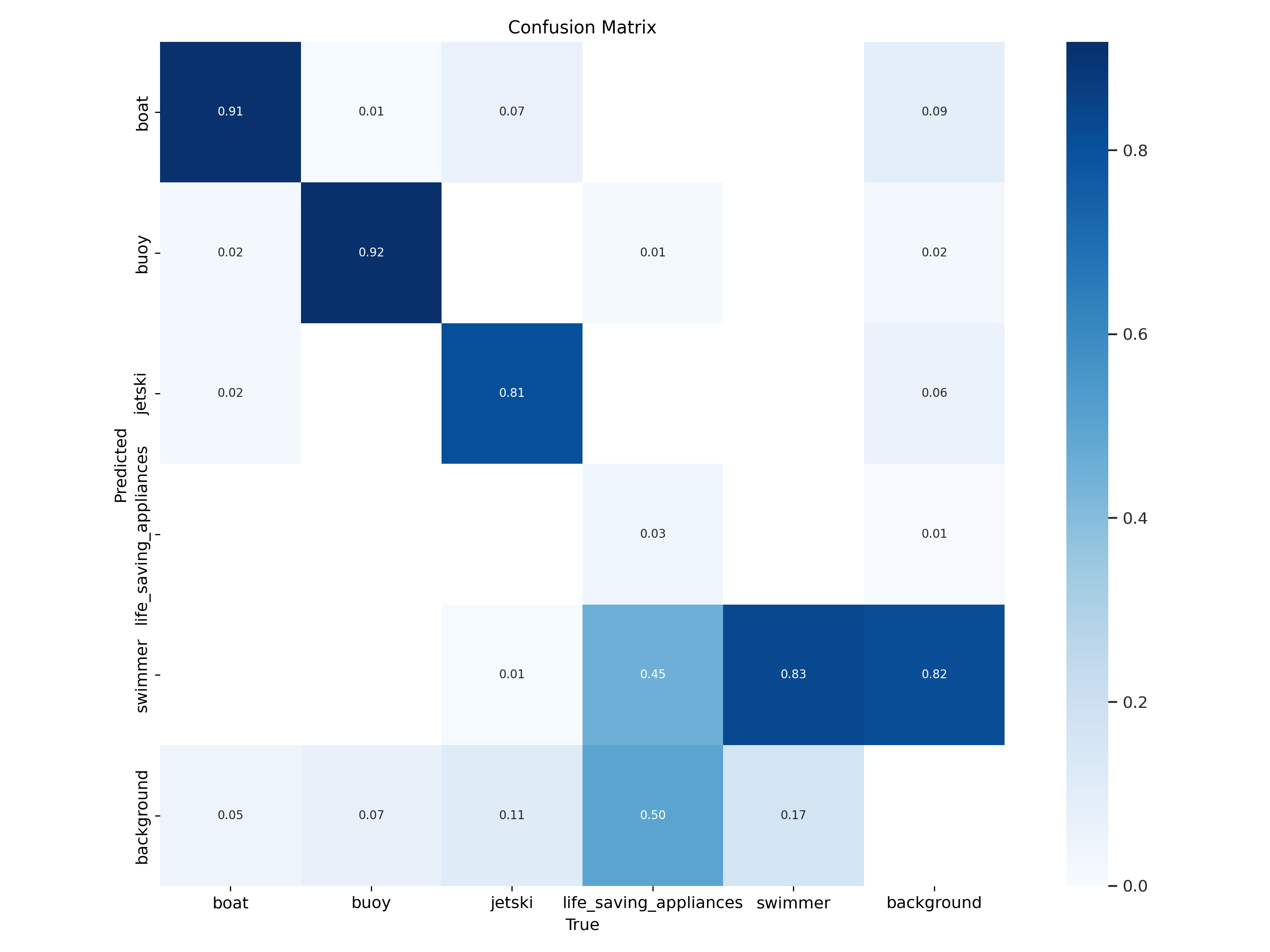}
        \caption{Confusion Matrix (training)}
        \label{sfig:yolotrainconfusion}
    \end{subfigure}
    \hfill
    \begin{subfigure}[b]{0.48\textwidth}
        \centering
        \includegraphics[width=.75\textwidth]{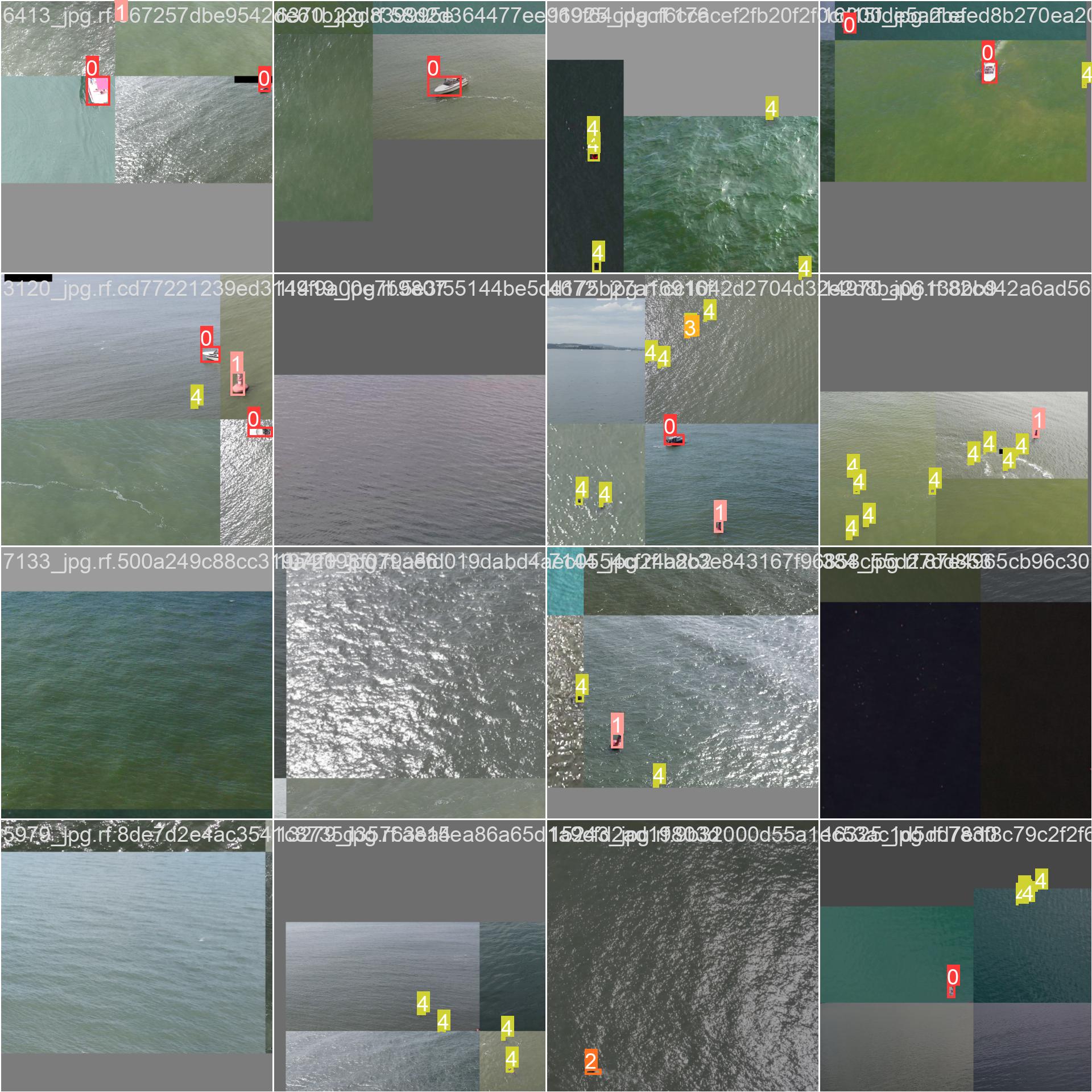}
        \caption{Training-time Mosaic augmentation (16 samples from training batch 1)}
        \label{sfig:yolotrainmosaicaug}
    \end{subfigure}

    \caption{YOLOv5s training statistics (200 epochs). Validation set performance: $mAP_{50}=0.69$, $mAP_{0.5:0.95}=0.45$.}
    \label{fig:yolov5training}
\end{figure*}

We trained the pre-trained YOLOv5s model for 200 epochs using the default configuration at 640 pixels resolution (image's longest side resized at 640p keeping aspect ratio).
Since our training datasets are 2x2 tiled, essentially, we trained the model at 1280p resolution of the input images(regardless of their source) and 2x scale of their original pixel-print in the source images. Although tiled training does not directly translate to resolution scaling, it enables the model to see much finer details at the training resolution(640) which would have been lost due to down-sampling, i.e., ${5456p}\times{3632p}$ images from 200+meters altitude, down-sampled to ${640p}\times{426p}$ will not have any discernible features to detect the swimmers.
We did not strictly scale the training sets, as YOLOv5 is to some extent scale invariant by design, thanks to its normalized auto-anchor algorithm and mosaic augmentation. \Cref{sfig:yolotrainmosaicaug} shows how the training input is augmented into mosaics of random scales. The authors claim this helps the model to learn to recognize the same objects at different scales \cite{yolov5x_mosaic}.

\textbf{Training statistics:}
\Cref{fig:yolov5training} shows the training matrices over 200 epochs. The F1-confidence graph in \Cref{sfig:yolotrainf1} looks good for all classes except \say{life-saving-appliances}, and we will discuss it further in the discussion section. However, if we look at the confusion matrix \Cref{sfig:yolotrainconfusion}, we see that our model is also detecting a lot of false positives for \say{swimmer} class, one of the smallest classes in terms of object size. The consequences of this will be apparent in \Cref{sec:yolo_res}.


\section{Results and Discussion}
\label{sec:yolo_res}

We evaluated our altitude-aware dynamic tiling approach on the \textit{SeaDronesSee} test images using the two test sets described above (\textit{ScaledV1} and \textit{ScaledV2}). We established two baselines:
\begin{enumerate}
    \item using \textbf{no tiling}, i.e., running YOLOv5 on the original images resized to 640px (default input size);
    and
    \item using \textbf{static tiling} with \textit{\ac{SAHI}}, i.e., slicing each image into 640px tiles (with 20\% overlap) but without any altitude-based scaling adjustment.
\end{enumerate}
The performance of each configuration is summarized in \Cref{tab:results0} in terms of detection accuracy and speed. We report the \ac{mAP} at various IoU thresholds and object size categories and the average inference speed in \ac{FPS}.

\begin{table*}[hbtp!]
\caption{mAP and \ac{FPS} comparisons across various test setups}
\label{tab:results0}
\resizebox{\textwidth}{!}{%
\begin{tabular}{lrrrrrr}
\hline
\multicolumn{1}{l}{\textbf{Test Setup}} & \multicolumn{1}{l}{\textbf{mAP}} & \multicolumn{1}{l}{\textbf{mAP50}} & \multicolumn{1}{c}{\textbf{mAP50}} & \multicolumn{1}{c}{\textbf{mAP50}} & \multicolumn{1}{c}{\textbf{mAP50}} & \multicolumn{1}{c}{\textbf{FPS}} \\ 
\multicolumn{1}{l}{} & \multicolumn{1}{l}{} & \multicolumn{1}{l}{} & \multicolumn{1}{r}{small} & \multicolumn{1}{r}{medium} & \multicolumn{1}{r}{large} & \multicolumn{1}{c}{(avg)} \\ \hline
Original image (640px) & {\color[HTML]{CB0000} 0.16} & {\color[HTML]{CB0000} 0.35} & 0.34 & 0.88 & 1.00 & \textbf{7.00} \\
Tiled image (SAHI) & \textbf{0.34} & \textbf{0.62} & 0.42 & 0.75 & 0.91 & {\color[HTML]{CB0000} 0.60} \\
\textbf{ScaledV1 (ours)} & 0.25 & 0.48 & {\color[HTML]{009901} \textbf{0.45}} & 0.68 & 0.97 & {\color[HTML]{009901} \textbf{2.60}} \\
\textbf{ScaledV2 (ours)} & 0.27 & {\color[HTML]{009901} 0.52} & {\color[HTML]{009901} \textbf{0.47}} & 0.73 & 0.97 & 1.87 \\  \hline
\end{tabular}
}

\end{table*}
\subsection{Baseline vs. Tiling}

Using the standard COCO-style \ac{mAP} (averaged over IoU thresholds 0.5:0.95 for all classes), the baseline scored a low \ac{mAP} of 0.16, highlighting the challenge of detecting small objects with resource constraints (downscaling 640px).
Employing static \ac{SAHI} tiling markedly improves the overall performance to \ac{mAP}50 = 0.62, significantly boosting detection accuracy for small objects ($mAP50_{small} = 0.42$) but slightly reducing performance on medium and large objects (0.75 and 0.91, respectively).
However, this accuracy gain comes at a severe cost to inference speed, reducing it to an impractical 0.60 \ac{FPS}.

\subsection{Proposed Dynamic Approach (ScaledV1 \& ScaledV2)}
On the other hand, our altitude-aware dynamic tiling approaches, ScaledV1 and ScaledV2, strike a more effective balance between accuracy and inference speed, especially for small objects, while maintaining significantly higher speeds than the static tiled approach.

ScaledV2 reached an overall \ac{mAP} of 0.27 (a 69\% increase over the 0.16 non-tiled baseline) and a small-object mAP50 of 0.47 (up from 0.34 baseline). These accuracy figures, although slightly lower than the static \ac{SAHI} tiling’s scores, were attained at $3\times$ to $4\times$ faster inference speed (1.87–2.60 \ac{FPS} vs 0.60 \ac{FPS}). Notably, our method even surpassed the static tiling in the specific metric of small-object precision: both ScaledV1 and ScaledV2 achieved higher mAP50 for small objects (0.45 and 0.47, respectively) than the static tiled baseline’s 0.42. This indicates that altitude-aware scaling successfully boosts the visibility of small objects to the detector, but without having to tile the image as extensively. At the same time, the \ac{mAP} for large objects remained close to 1.0 for all methods.
Medium-sized object detection showed a slight decrease in our scaled setups (e.g. mAP50 (medium) of 0.68–0.73 for ScaledV1/V2 vs 0.88 in the 640px baseline). This suggests that the focus on small-object enhancement did not significantly harm large-object detection but did trade a bit of the mid-scale accuracy. In practice, this trade-off should be acceptable because medium and large objects were already being detected with high confidence by the base model, and the primary challenge was the small objects.

Between our two proposed configurations, ScaledV1 vs. ScaledV2, we see the effect of a more aggressive scaling strategy. ScaledV2 was designed to use finer-grained (more aggressive) altitude-based scaling effectively, resulting in more tiles (higher resolution) for a given high-altitude image than ScaledV1. Consequently, ScaledV2 achieved slightly higher accuracy across the board (e.g., overall \ac{mAP} 0.27 vs 0.25, and small-object mAP50 0.47 vs 0.45) compared to ScaledV1. However, this came with a reduction in speed from 2.60 \ac{FPS} (ScaledV1) to 1.87 \ac{FPS} (ScaledV2). In other words, ScaledV2 improves small-object mAP50 by 2\% over ScaledV1 but reduces inference speed by 28\%, highlighting a tunable accuracy-speed trade-off.


\begin{figure*}[htbp!]
    \centering
    \begin{subfigure}[b]{0.32\textwidth} 
        \centering
        \includegraphics[width=\textwidth]{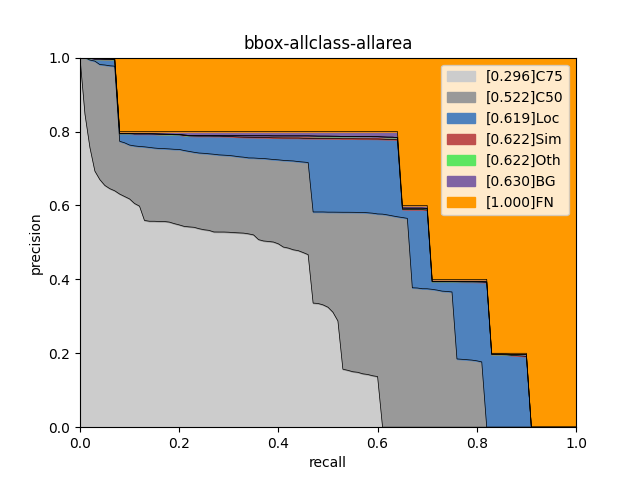}
    \end{subfigure}
    \hfill
    \begin{subfigure}[b]{0.32\textwidth} 
        \centering
        \includegraphics[width=\textwidth]{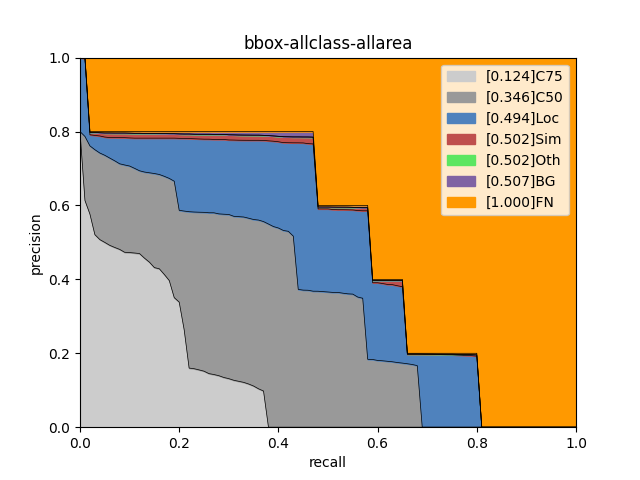}
    \end{subfigure}
    \hfill
    \begin{subfigure}[b]{0.32\textwidth} 
        \centering
        \includegraphics[width=\textwidth]{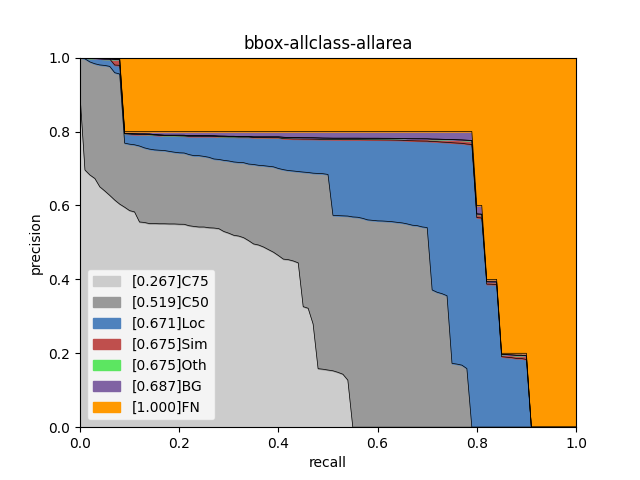}
    \end{subfigure}

    \medskip 
    
    \begin{subfigure}[b]{0.32\textwidth} 
        \centering
        \includegraphics[width=\textwidth]{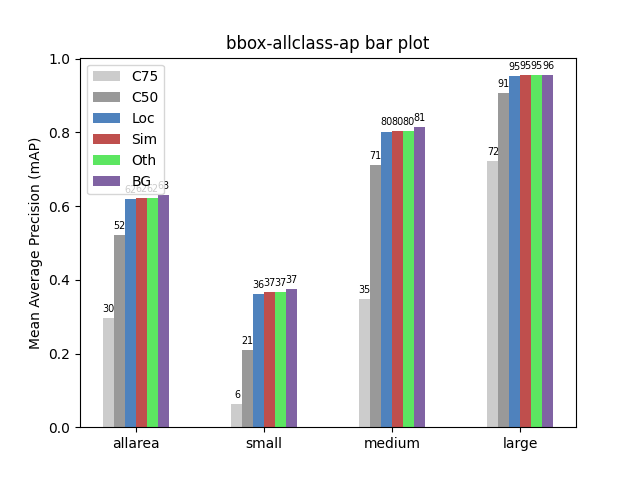}
        \caption{Original full resolution}
    \end{subfigure}
    \hfill
    \begin{subfigure}[b]{0.32\textwidth} 
        \centering
        \includegraphics[width=\textwidth]{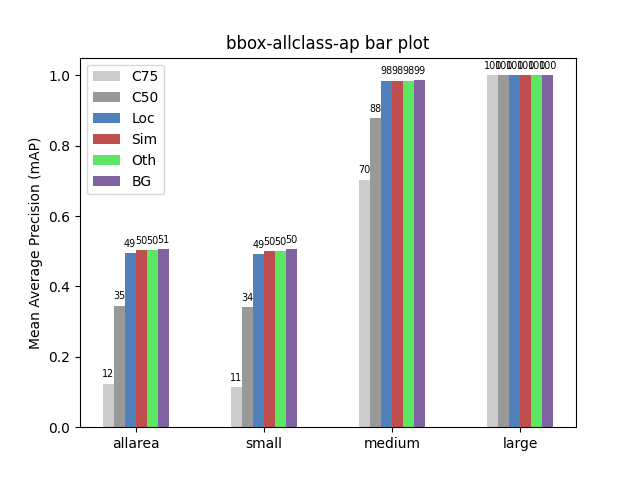}
        \caption{Original resized @640px}
    \end{subfigure}
    \hfill
    \begin{subfigure}[b]{0.32\textwidth} 
        \centering
        \includegraphics[width=\textwidth]{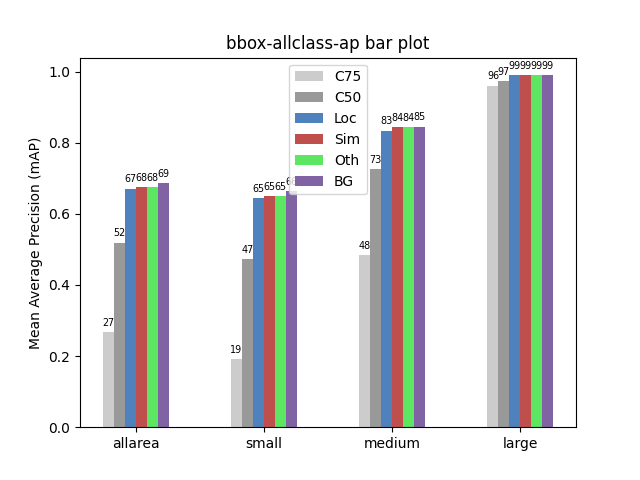}
        \caption{ScaledV2: \ac{DSS}+\ac{SAHI} (\textbf{ours})}
    \end{subfigure}
    
    \caption{Test set performance matrices. The top row shows bbox metrics for all classes and areas, bottom row shows AP bar plots. Columns represent Original images at full resolution (left), Original images resized @640px (middle), and \textit{our proposed method} (ScaledV2: DSS+SAHI @640px) (right).}
    \label{fig:test_results_yolo}
\end{figure*}


It should be noted that our YOLOv5s detector was used with default training settings, without extensive hyperparameter tuning specifically for the SeaDronesSee dataset. Consequently, absolute \ac{mAP} values for challenging categories, such as swimmers or small floating life-saving appliances, remained relatively modest.
Even when we removed the input size restriction(pass full-resolution image to the YOLOv5 detector), which is unlikely in a real-world scenario, it still loses to \ac{DSS}+\ac{SAHI} for small object detection ($mAP50_{small}$) 0.21 for full resolution and 0.47 for ScaledV2) (see \Cref{fig:test_results_yolo}), reflecting the inherent difficulty of these models. Nonetheless, our altitude-aware dynamic tiling consistently improved detection accuracy across all object sizes. Compared to the no-tiling baseline, we achieved approximately 38\% higher accuracy ($mAP50_{small}$) for critical small objects, while simultaneously more than tripling inference speed compared to static exhaustive tiling. This demonstrates that dynamically adjusting image resolution and tiling density with altitude effectively balances accuracy and computational efficiency, crucial for practical maritime UAV \ac{SAR} operations

\section{Conclusion and Future Work}

In this paper, we introduced an altitude-aware dynamic tiling approach that adapts image scaling and tiling factors to enhance small object detection performance in maritime UAV imagery. This could be particularly important for SAR scenarios. By coupling the tiling strategy directly with flight altitude and object scale, our method achieves significantly higher accuracy and efficiency compared to baseline approaches such as no tiling or static tiling. Experimental results on the SeaDronesSee dataset demonstrated substantial improvements, including 38\% higher small-object detection accuracy  ($mAP50_{small}$) relative to a no-tiling baseline and more than threefold inference speed improvement over static tiling.
Crucially, these enhancements were achieved without modifying the underlying detector architecture. This underscores how altitude-aware dynamic tiling (DSS+SAHI) can effectively complement existing deep-learning models in computationally constrained environments.

The first natural continuation of this work would be further developing the architecture and conducting field experiments with \acp{SBC}.
Deploying our pipeline in real-world operations will test its robustness and reveal necessary optimizations for handling dynamic flight conditions.

Secondly, in this experiment, we only used one model (YOLOv5s) to maintain consistency and comparability. As our proposed framework is architecture agnostic, there is considerable potential to integrate and evaluate \ac{SOTA} vision libraries and detection architectures, or develop a standalone platform like \ac{SAHI}.
Specialized YOLO variants optimized for small-object detection (e.g., SOD-YOLO, HIC-YOLOv5) \cite{SOD-YOLO, HIC-YOLOv5} could be experimented with to further reduce the computational burden or increase speed and accuracy.

Finally, extending our approach to leverage multi-modal data, such as combined thermal and visible imaging (\ac{RGBT}), hyperspectral images could improve detection performance by utilizing complementary sensor strengths, especially in challenging visibility conditions.

In summary, our altitude-aware dynamic tiling method effectively bridges the gap between high detection accuracy and practical real-time performance in challenging maritime contexts for small objects. We look forward to exploring these research avenues to further enhance the robustness and applicability of our method in operational UAV deployments.

\section*{Acknowledgment}
This research was developed as part of an MSc thesis in the Erasmus Mundus Master in Marine and Maritime Intelligent Robotics (MIR) program at the Department of Marine Technology, Norwegian University of Science and Technology.


\bibliographystyle{IEEEtran}
\bibliography{IEEEabrv,thesis}

\begin{thebibliography}{10}
\providecommand{\url}[1]{#1}
\csname url@samestyle\endcsname
\providecommand{\newblock}{\relax}
\providecommand{\bibinfo}[2]{#2}
\providecommand{\BIBentrySTDinterwordspacing}{\spaceskip=0pt\relax}
\providecommand{\BIBentryALTinterwordstretchfactor}{4}
\providecommand{\BIBentryALTinterwordspacing}{\spaceskip=\fontdimen2\font plus
\BIBentryALTinterwordstretchfactor\fontdimen3\font minus \fontdimen4\font\relax}
\providecommand{\BIBforeignlanguage}[2]{{%
\expandafter\ifx\csname l@#1\endcsname\relax
\typeout{** WARNING: IEEEtran.bst: No hyphenation pattern has been}%
\typeout{** loaded for the language `#1'. Using the pattern for}%
\typeout{** the default language instead.}%
\else
\language=\csname l@#1\endcsname
\fi
#2}}
\providecommand{\BIBdecl}{\relax}
\BIBdecl

\bibitem{dataset_SDSv2}
L.~A. Varga, B.~Kiefer, M.~Messmer, and A.~Zell, ``Seadronessee: A maritime benchmark for detecting humans in open water,'' in \emph{2022 IEEE/CVF Winter Conference on Applications of Computer Vision (WACV)}, 2022, pp. 3686--3696.

\bibitem{yolov5}
Ultralytics, ``{YOLOv5}: {A} state-of-the-art real-time object detection system,'' \url{https://docs.ultralytics.com}, 2021, accessed: 2024-12-20.

\bibitem{sahi_app}
\BIBentryALTinterwordspacing
F.~C. Akyon, C.~Cengiz, S.~O. Altinuc, D.~Cavusoglu, K.~Sahin, and O.~Eryuksel, ``Sahi: A lightweight vision library for performing large scale object detection and instance segmentation,'' 11 2021. [Online]. Available: \url{https://doi.org/10.5281/zenodo.5718950}
\BIBentrySTDinterwordspacing

\bibitem{smol_object_anchorbox}
\BIBentryALTinterwordspacing
E.~Mohamed, K.~Sirlantzis, G.~Howells, and S.~Hoque, ``Optimisation of deep learning small-object detectors with novel explainable verification,'' \emph{Sensors}, vol.~22, no.~15, 2022. [Online]. Available: \url{https://www.mdpi.com/1424-8220/22/15/5596}
\BIBentrySTDinterwordspacing

\bibitem{tiling_power}
F.~O. \"{U}nel, B.~O. \"{O}zkalayci, and C.~\c{C}i\u{g}la, ``The power of tiling for small object detection,'' in \emph{2019 IEEE/CVF Conference on Computer Vision and Pattern Recognition Workshops (CVPRW)}, 2019, pp. 582--591.

\bibitem{sahi}
F.~C. Akyon, S.~O. Altinuc, and A.~Temizel, ``Slicing aided hyper inference and fine-tuning for small object detection,'' \emph{2022 IEEE International Conference on Image Processing (ICIP)}, pp. 966--970, 2022.

\bibitem{Zolich2021-tk}
A.~Zolich, T.~A. Johansen, M.~Elkolali, A.~Al-Tawil, and A.~Alcocer, ``Unmanned aerial system for deployment and recovery of research equipment at sea,'' in \emph{2021 Aerial Robotic Systems Physically Interacting with the Environment ({AIRPHARO})}, 2021, pp. 1--8.

\bibitem{Leira2021-mo}
F.~S. Leira, H.~H. Helgesen, T.~A. Johansen, and T.~I. Fossen, ``\BIBforeignlanguage{en}{Object detection, recognition, and tracking from {UAVs} using a thermal camera},'' \emph{\BIBforeignlanguage{en}{J. field robot.}}, vol.~38, no.~2, pp. 242--267, 2021.

\bibitem{Leira_2015_detect}
F.~S. Leira, T.~A. Johansen, and T.~I. Fossen, ``Automatic detection, classification and tracking of objects in the ocean surface from uavs using a thermal camera,'' in \emph{2015 IEEE Aerospace Conference}.\hskip 1em plus 0.5em minus 0.4em\relax IEEE, 2015.

\bibitem{Leira_2015_georef}
F.~S. Leira, K.~Trnka, T.~I. Fossen, and T.~A. Johansen, ``A ligth-weight thermal camera payload with georeferencing capabilities for small fixed-wing uavs,'' in \emph{2015 International Conference on Unmanned Aircraft Systems (ICUAS)}.\hskip 1em plus 0.5em minus 0.4em\relax IEEE, 2015.

\bibitem{Li_Wu_Kittler_2018}
H.~Li, X.-J. Wu, and J.~Kittler, ``Infrared and visible image fusion using a deep learning framework,'' in \emph{2018 24th International Conference on Pattern Recognition (ICPR)}.\hskip 1em plus 0.5em minus 0.4em\relax IEEE, 2018, pp. 2705--2710.

\bibitem{Kumar2022ImprovedYOLO3}
N.~Kumar, A.~K. Jilani, P.~Kumar, and A.~Nikiforova, ``Improved yolov3-tiny object detector with dilated cnn for drone-captured images,'' \emph{2022 International Conference on Intelligent Data Science Technologies and Applications (IDSTA)}, pp. 89--94, 2022.

\bibitem{mfgnet_rgbt}
X.~Wang, X.~Shu, S.~Zhang, B.~Jiang, Y.~Wang, Y.~Tian, and F.~Wu, ``Mfgnet: Dynamic modality-aware filter generation for rgb-t tracking,'' \emph{IEEE Transactions on Multimedia}, 2022.

\bibitem{rgbt-alignment}
Z.~Tu, Z.~Li, C.~Li, and J.~Tang, ``Weakly alignment-free rgbt salient object detection with deep correlation network,'' \emph{IEEE Transactions on Image Processing}, vol.~31, pp. 3752--3764, 2022.

\bibitem{Drone_rgbir}
\BIBentryALTinterwordspacing
L.~Yang, R.~Ma, and A.~Zakhor, ``Drone object detection using rgb/ir fusion,'' 2022. [Online]. Available: \url{https://arxiv.org/abs/2201.03786}
\BIBentrySTDinterwordspacing

\bibitem{rgbt_salient}
\BIBentryALTinterwordspacing
Z.~Tu, Y.~Ma, Z.~Li, C.~Li, J.~Xu, and Y.~Liu, ``Rgbt salient object detection: A large-scale dataset and benchmark,'' 2020. [Online]. Available: \url{https://arxiv.org/abs/2007.03262}
\BIBentrySTDinterwordspacing

\bibitem{oscar_physics_scaling}
J.~Walker, T.~Yamada, A.~Prugel-Bennett, and B.~Thornton, ``The effect of physics-based corrections and data augmentation on transfer learning for segmentation of benthic imagery,'' in \emph{2019 IEEE Underwater Technology (UT)}, 2019, pp. 1--8.

\bibitem{holistic_drone}
P.~Petrides, C.~Kyrkou, P.~Kolios, T.~Theocharides, and C.~Panayiotou, ``Towards a holistic performance evaluation framework for drone-based object detection,'' in \emph{2017 International Conference on Unmanned Aircraft Systems (ICUAS)}, 2017, pp. 1785--1793.

\bibitem{Raster_Tiling}
\BIBentryALTinterwordspacing
 [Online]. Available: \url{https://up42.com/marketplace/blocks/processing/tiling}
\BIBentrySTDinterwordspacing

\bibitem{liu2016ssd}
W.~Liu, D.~Anguelov, D.~Erhan, C.~Szegedy, S.~Reed, C.-Y. Fu, and A.~C. Berg, ``{SSD}: Single shot multibox detector,'' in \emph{ECCV}, 2016.

\bibitem{cocodataset}
\BIBentryALTinterwordspacing
T.~Lin, M.~Maire, S.~J. Belongie, L.~D. Bourdev, R.~B. Girshick, J.~Hays, P.~Perona, D.~Ramanan, P.~Doll{'{a} }r, and C.~L. Zitnick, ``Microsoft {COCO:} common objects in context,'' \emph{CoRR}, vol. abs/1405.0312, 2014. [Online]. Available: \url{http://arxiv.org/abs/1405.0312}
\BIBentrySTDinterwordspacing

\bibitem{camera_dist}
P.~J.~A. Alphonse and K.~V. Sriharsha, ``Depth perception in single rgb camera system using lens aperture and object size: a geometrical approach for depth estimation,'' \emph{SN Applied Sciences}, vol.~3, no.~6, p. 595, 5 2021.

\bibitem{roboflow}
\BIBentryALTinterwordspacing
B.~Dwyer, J.~Nelson, and J.~Solawetz, ``Roboflow,'' 2022. [Online]. Available: \url{https://roboflow.com}
\BIBentrySTDinterwordspacing

\bibitem{yolov5x_mosaic}
F.~Dadboud, V.~Patel, V.~Mehta, M.~Bolic, and I.~Mantegh, ``Single-stage uav detection and classification with yolov5: Mosaic data augmentation and panet,'' in \emph{2021 17th IEEE International Conference on Advanced Video and Signal Based Surveillance (AVSS)}, 2021, pp. 1--8.

\bibitem{SOD-YOLO}
Y.~Xiao and N.~Di, ``\BIBforeignlanguage{en}{{SOD-YOLO}: A lightweight small object detection framework},'' \emph{\BIBforeignlanguage{en}{Sci. Rep.}}, vol.~14, no.~1, p. 25624, 2024.

\bibitem{HIC-YOLOv5}
\BIBentryALTinterwordspacing
S.~Tang, S.~Zhang, and Y.~Fang, ``Hic-yolov5: Improved yolov5 for small object detection,'' 2023. [Online]. Available: \url{https://arxiv.org/abs/2309.16393}
\BIBentrySTDinterwordspacing

\end{thebibliography}


\end{document}